\documentclass[runningheads]{llncs}
\usepackage[T1]{fontenc}
\usepackage{graphicx}
\usepackage{amsmath}
\usepackage{amsfonts}
\usepackage{hyperref}
\usepackage[dvipsnames]{xcolor}
\usepackage{amssymb}

\usepackage{booktabs}
\usepackage{longtable}
\usepackage{makecell}
\usepackage{amsmath}

\usepackage{algorithm}
\usepackage{algpseudocode}

\begin{document}
\title{Artificial Neural Networks as Surrogate Models in Black Box Optimization\thanks{A version of this work has been accepted for publication at the European Conference on Machine Learning and Principles and Practice of Knowledge Discovery in Databases (ECML PKDD 2026).}}
\titlerunning{Black Box Optimization}
%
\author{Md Khadimul Islam Zim\inst{1} \and
Martin Hole\v{n}a\inst{1}}
\authorrunning{M.K.I. Zim and M. Hole\v{n}a}
%
\institute{Czech Academy of Sciences, Institute of Computer Science,\\
Pod Vodárenskou v\v{e}\v{z}í 2, Prague, Czech Republic\\
\email{mekuzim@gmail.com, martin@cs.cas.cz}}
%
\maketitle 
%

\begin{abstract}
Black-Box Optimization (BBO) is often applied in several engineering fields and can utilize an advancement of numerical measurements and simulation technologies. It deals with the optimization functions, where an analytical description is unavailable. It relies on methods that require only an input point in the search space, paired with its corresponding objective function value, obtained through non-analytical means, e.g., sensors, experiments, or simulations. Common approaches include evolutionary optimization and other metaheuristics. Since BBO methods rely solely on objective function values, they typically require many evaluations, which becomes problematic when evaluating the objective function is time-consuming or expensive. This leads to using surrogate-based optimization which evaluates selected true objective values and trains a regression model to approximate the objective function across the search space. Surrogate-assisted black-box optimization is a small-data learning problem because the optimizer must approximate an expensive objective function from limited evaluations. Surrogate models act as data-efficient regressors, guiding the search toward promising or informative points under a restricted evaluation budget. In this paper, a new surrogate model using artificial neural networks, called Adaptive-Fidelity Nexus Covariance Matrix Adaptation Evolution Strategy (AFN-CMA-ES), is proposed for the selective evaluation of objective functions. The experimental results show its competitive performance compared to state-of-the-art surrogate-assisted BBO methods.

\end{abstract}
\section{Introduction}
Black-box optimization (BBO) is concerned with optimization problems where for the objective function, an expression as explicit or implicit analytical expression is not available. In such problems, the optimization algorithm can only evaluate candidate solutions and observe the corresponding objective function values. These problems are often encountered in engineering design, where the performance of a system can only be assessed by performing numerical simulations, physical experiments, sensor readings or other non-analytical methods.

BBO is important because many scientific and engineering optimization problems involve objective functions that are not available analytically and are typically evaluated through experiments or costly simulations. Because such evaluations are expensive, the evaluation budget is usually limited, and the main challenge is to find good solutions with as few true objective-function evaluations as possible. Among state-of-the-art methods for continuous BBO, the Covariance Matrix Adaptation Evolution Strategy (CMA-ES) is widely used for nonlinear and non-convex problems because it adapts the covariance matrix of the sampling distribution and can learn dependencies among decision variables during the search process \cite{Hansen2003CMAES}. CMA-ES samples candidate solutions, ranks them, and updates its search distribution toward promising regions. Recent benchmark studies also suggest the need to evaluate BBO algorithms on more realistic problem classes, such as mixed-integer optimization problems \cite{marty2023mixedinteger}.

Since BBO methods rely solely on objective function values, common approaches such as evolutionary algorithms and other metaheuristics often require many evaluations. This is less efficient when the evaluations are slow or expensive. Surrogate-assisted BBO overcomes this problem by limiting the number of candidate solutions evaluated by the objective function and by learning a regression model called surrogate model that estimates the objective function over the solution space. The surrogate then guides the optimizer toward promising or informative regions while reducing the number of costly  evaluations \cite{bajer2015benchmarking}.

Surrogate-assisted BBO is important from the small-data perspective because it evaluates the true objective function only for selected candidates and uses these evaluated input–output pairs to train a surrogate model that guides the search. Thus, in many machine-learning applications, the surrogate model in BBO cannot be trained on a large dataset because each training sample requires a costly true objective-function evaluation. Rather, the optimizer must learn a good approximation to the objective function from a small number of costly input-output pairs. As a result, accurate prediction, uncertainty quantification, and adaptive sampling are key to surrogate-assisted BBO. Various types of surrogate models, such as Gaussian processes, polynomials, random forests and other regression models, are used to assist the optimizer with minimal function calls.

\subsection{Aim}
The goal of the reported research is to develop and validate the Adaptive-Fidelity Nexus (AFN). This new surrogate model helps efficiently optimize computationally expensive functions using an adaptively refined artificial neural network (ANN). The main goal is to develop a smart refinement, which will be carefully tested using several key measures.
The main research objectives are:
\begin{itemize}
    \item To create the adaptive AFN, which combines ANNs as surrogate models with intelligent sampling methods based on uncertainty quantification.
\end{itemize}
\begin{itemize}
    \item To substantially reduce the amount of costly objective function evaluations while getting sufficiently close to the optimal solution.
\end{itemize}
\begin{itemize}
    \item To investigate the dependence of the AFN performance on the character of the benchmark function.
\end{itemize}
\begin{itemize}
    \item To enhance the robustness and resource efficiency of the optimization process for a range of problem instances.
\end{itemize}
\section{Literature Review}
Surrogate-assisted optimization has grown rapidly as an effective approach to BBO problems that need to keep low the number of true function evaluations \cite{bajer2015benchmarking}. Research performed so far demonstrates that surrogate-assisted CMA-ES can reduce the number of expensive fitness evaluations. Gaussian processes, random forests, and local quadratic models have been extensively investigated as surrogate models for CMA-ES, and their performance depends on the problem landscape, dimensionality, and evolution-control strategy \cite{bajer2019gp}. Ordinal Gaussian process regression have also been investigated for surrogate-assisted CMA-ES, due to CMA-ES invariance with respect to monotonously increasing transformations \cite{pitra2017ordinal}. Recently, modern neural-network-based surrogate models have been investigated for active and transfer learning in surrogate-assisted black-box optimization. However, they also present issues related to uncertainty estimation, stability, and scalability \cite{holena2024modern}. AFN-CMA-ES extends neural-network surrogate-assisted BBO by integrating a bootstrapped  multilayer perceptron (MLP) ensemble, rank-normalized local-archive training, anchored regularization, uncertainty estimation, and Kendall's $\tau$-based adaptive model life within CMA-ES.

\section{Proposed Methodology}
\subsection{Overview}
The proposed AFN-CMA-ES framework extends the standard CMA-ES by integrating it with the Adaptive-Fidelity Nexus (AFN) surrogate system. Rather than using an expensive function in every iteration, AFN-CMA-ES uses a trained ANN ensemble to predict values and estimate their uncertainty. It is called Adaptive-Fidelity Nexus because it adaptively combines true evaluations, surrogate predictions, and CMA-ES within one framework. AFN-CMA-ES was implemented according to the proposed framework, using the ANN ensemble surrogate, generation-based evolution control, and CMA-ES update procedure summarized in Algorithm \ref{alg:afn_cmaes} below.
\subsection{Adaptive-Fidelity Nexus}
AFN operates in $3$ main stages:
\begin{enumerate}
    \item Surrogate Training: An ensemble of $M = 5$ MLP networks is trained on the current evaluation samples. Although all ensemble members use the same MLP architecture, prediction diversity is ensured through independent training. For a local archive of \(N_{\mathrm{loc}}\) samples, each bootstrap sample contains \(N_{\mathrm{loc}}\) observations drawn with replacement. The level of disagreement among them is used to estimate the epistemic uncertainty. There is a prediction $\hat{f}(x)$ within each network, and the variance estimates the uncertainty.

\algrenewcommand\algorithmicindent{1.2em}
\begin{algorithm}[H]
\caption{AFN-CMA-ES }
\label{alg:afn_cmaes}
\begin{algorithmic}[1]
\Require Objective function $f(x)$, bounds $[lb,ub]$, budget $B$, ensemble size $M$, model life $L$
\Ensure Best evaluated solution $x^{*}$

\State Generate initial Latin hypercube sampling (LHS) archive $D=\{(x_i,f(x_i))\}_{i=1}^{N_0}$, where $N_0=20d$.
\State Train an ensemble of $M$ ANN surrogates on rank-normalized objective values.
\State Initialize CMA-ES from the best sample in $D$ and set $t\leftarrow N_0$.

\While{$t<B$}
    \State Generate CMA-ES population $X_g=\{x_{g,1},\ldots,x_{g,\lambda}\}$.
    \State Evaluate all candidates with the true objective function and update $D$, $x^{*}$, and CMA-ES.
    \State Estimate surrogate ranking quality $\tau_g$ using Kendall's correlation.
    \Statex \hspace{\algorithmicindent}\textit{$\tau_g$ is one scalar value for generation $g$ and is computed before retraining to assess the surrogate model used in the previous surrogate-only generation.}
    \State Retrain the ANN ensemble on a local archive subset around the CMA-ES mean.
    \State Set model life $L_g=L$ in fixed mode, or choose $L_g$ according to $\tau_g$ in adaptive mode.
    \For{$j=1$ to $L_g$}
        \State Generate surrogate-only CMA-ES population $\widetilde{X}_{g,j}$.
        \State Predict values using ensemble mean $\hat{\mu}(x)$.
        \State Update CMA-ES using rank-normalized surrogate fitness scores.
    \EndFor
    \State Update $t\leftarrow |D|$.
\EndWhile

\State \Return $x^{*}$ and $f(x^{*})$
\end{algorithmic}
\end{algorithm}

\item \text{Surrogate only ranking:} Training uses the initial LHS archive and the local archive subset selected around the current CMA-ES mean, where $d$ is the problem dimension. This dimension aware design prevents the MLP from fitting on insufficient data in higher dimensional spaces. The ensemble computes:
\begin{itemize}
\item[] Predictive mean (Exploitation):
\item[] $\hat{\mu}(x) = \frac{1}{M} \sum_{i=1}^{M} \hat{f}_i(x)$ 
\item[] Predictive variance (Exploration):
\item[] $\hat{\sigma}^2(x) = \frac{1}{M} \sum_{i=1}^{M} \left( \hat{f}_i(x) - \hat{\mu}(x) \right)^2$
\end{itemize}
The ensemble predictive mean is used to rank candidate solutions during surrogate only CMA-ES generations. The ensemble variance provides uncertainty information, but the revised implementation does not use an individual uncertainty threshold to decide whether each candidate should be evaluated by the true objective function. Instead, true evaluations are controlled at the generation level.

\item Iterative Refinement: New evaluations are added to the archive, and a local archive subset of at most 250 samples is selected according to distance from the current CMA-ES mean. This limit balances local relevance and retraining cost. In the initial phase, when the number of available evaluations is less than $250$, all available samples are used. The surrogate is retrained using this local archive subset after each true evaluated generation. This helps the surrogate remain focused on the current search region, since samples far from the current estimate of the optimum may be more misleading than older samples near the optimum.

\end{enumerate}
\subsection{Integration with CMA-ES}
In AFN-CMA-ES, the surrogate is used instead of costly black-box function evaluations within the evolution control loop. CMA-ES is used to generate candidate solutions from its sampling distribution. AFN-CMA-ES uses generation based evolution control. At each generation evaluating the true black-box function, CMA-ES generates a population of $\lambda$ candidate solutions, where $\lambda$ denotes the population size, and all candidates are evaluated using the true objective function. These evaluated samples are added to the archive and are used to update both the CMA-ES distribution and the ANN ensemble surrogate.

 The model life $L_g$ denotes the number of consecutive surrogate-only CMA-ES generations performed after the generation g evaluating the true black-box function. During these surrogate only generations, CMA-ES again generates populations of $\lambda$ candidates, but their fitness values are estimated by the ANN ensemble. The ensemble predictive mean $\hat{\mu}(x)$ is used to rank the candidate solutions, and the predicted values are converted into rank normalized surrogate fitness scores. These surrogate ranked scores are then passed to CMA-ES for updating the mean, step size, and covariance matrix without spending additional true objective function evaluations.

This mechanism reduces the proportion of true evaluated CMA-ES generations while allowing additional CMA-ES adaptation steps under the same true evaluation budget. The model life $L$ is either fixed or adaptively selected according to Kendall's rank correlation between the previous surrogate predictions and the newly observed true objective values. When the surrogate ranking quality is high, more surrogate only generations are allowed; when the ranking quality is low, the model life is reduced.

Restart steps are used by AFN-CMA-ES when the CMA-ES internal stopping criterion is met or when stagnation is detected over true-evaluated generations. In such cases, the optimizer is reinitialized either near the best evaluated solution or from a random point in the search domain, depending on the restart count. This enables the algorithm to continue the search when premature convergence occurs. Then CMA-ES updates its covariance matrix by utilizing true evaluations during true-evaluated generations and surrogate ranked predictions during surrogate-only generations. This mechanism balances exploration and exploitation while reducing the number of costly true objective-function evaluations.

\subsection{Data Collection and Sampling}
This first step consists of creating a representative initial sample of the input-output pairs of the Comparing Continuous Optimizers (COCO) benchmark suite, denoted \( D_0 = \{ (x_i, y_i) \}_{i=1}^{N_0} \), from the original, costly objective function \( f(x) \). In this case, \( x_i \) is an input vector in the design space \( X \subseteq \mathbb{R}^d \) and \( y_i = f(x_i) \) is the scalar objective-function value. To ensure these \( N_0 \) data points are well distributed throughout the design space and capture its important features with minimal costly evaluations, in this study, LHS was used to generate the initial sample set because it provides space filling coverage of the design domain with a limited number of expensive function evaluations, the initial sample size was set to \(N_0=20d\), subject to the available evaluation budget. This dimension-scaled choice provides sufficient initial coverage while preserving most of the \(250d\) evaluation budget for optimization. Newly true evaluated CMA-ES samples were added to the archive during optimization.

\subsection{Data Pre-processing and Standardization}

Input/Output Standardization: Before training, the input features $X$ are standardized to a zero mean and unit standard deviation, and objective-function values $y$ are first converted into rank-normalized target values. These rank normalized target values are also standardized during neural network training. The predicted values are used to rank candidate solutions during surrogate only CMA-ES generations. This prevents the variables of greater magnitude from dominating the surrogate fit over the various scales of BBOB functions.

Regularized Ensemble Training: During surrogate retraining, the ANN ensemble is trained on the local archive subset selected around the current CMA-ES mean. Training uses standardized inputs and rank-normalized targets, with a loss function combining a negative-log-likelihood term and a mean-squared-error term, rank-normalized targets mean that the raw objective-function values are replaced by their ranks within the current training archive and then scaled before ANN training. Anchored regularization is applied by penalizing deviations from the initial network parameters. This helps avoid overfitting small archives at the beginning of the optimization run.

\subsection{Ensemble of ANN Surrogates}
The AFN framework uses an Ensemble of MLP networks as the surrogate models.
\begin{enumerate}
    \item MLP Architecture: 
    
\begin{itemize}
\item[]Input layer: $d$ neurons (problem dimension)
\item[]Hidden layer 1: 64 neurons with rectified linear unit (ReLU) activation
\item[]Hidden layer 2: 32 neurons with ReLU activation
\item[]Output layer: 1 neuron with linear activation
\end{itemize}

This compact 64--32 architecture balances nonlinear capacity, overfitting risk, and repeated retraining cost in the small-data setting.

\item Training Procedure:

The network is trained using the standardized rank-normalized target
values. Each network predicts both a mean value $\mu_i$ and a log-variance
value $\log \sigma_i^2$. The loss function combines a negative log-likelihood (NLL) term and a mean squared error (MSE) term:
\begin{equation}
L(W,B)=
\frac{1}{N}\sum_{i=1}^{N}
\frac{1}{2}
\left[
\log \sigma_i^2+
\frac{\left(y_i^{\mathrm{norm}}-\mu_i\right)^2}
{\exp\left(\log \sigma_i^2\right)}
\right]
+
\frac{1}{2N}\sum_{i=1}^{N}
\left(\mu_i-y_i^{\mathrm{norm}}\right)^2.
\end{equation}
where $W$ and $B$ denote the ANN weights and biases, respectively, and
$\mu_i-y_i^{\mathrm{norm}}$ is the prediction error in the standardized
rank-normalized target space for sample $i$.

The loss $L(W,B)$ combines NLL and MSE. NLL estimates each network's
predictive variance, while MSE stabilizes the mean prediction used for
candidate ranking. This variance complements ensemble disagreement,
which represents epistemic uncertainty across networks.

To prevent overfitting, anchored regularization is applied:
\begin{equation}
L_{\text{reg}}(W, B) = L(W, B) + \lambda_a \sum_{p \in W,B} \left(p-p_0\right)^2
\end{equation}

where \(L_{\text{reg}}(W,B)\) is the regularized loss, \(L(W,B)\) is the original loss, \(\lambda_a\) controls the strength of anchored regularization, \(p\) denotes a trainable network parameter, and \(p_0\) denotes its initial value.

Training uses the AdamW optimizer, which applies Adam with decoupled
weight decay. The basic gradient-based weight update can be written as:
\begin{equation}
w_{kj}^{\text{new}} = w_{kj}^{\text{old}} - \eta \frac{\partial L_{\text{reg}}}{\partial w_{kj}}
\end{equation}
where \(w_{kj}^{\text{new}}\) and \(w_{kj}^{\text{old}}\) are the updated and previous weights, respectively. Each network uses a bootstrap sample drawn from the local archive, and warm-starting reduces retraining cost while preserving learned information. The hyperparameters were selected to optimize for accuracy, stability, and cost.

\item Ensemble Construction:

Rather than a single MLP, we train $M = 5$ independent ANNs to form an Ensemble, which balances prediction diversity and computational cost.

For the query point $x$, the ensemble provides predictive mean and predictive variance (Exploration):
\begin{equation}
\hat{\sigma}^2(x) = \frac{1}{M} \sum_{i=1}^{M} \left[\hat{f}_i(x) - \hat{\mu}(x)\right]^2
\end{equation}
The variance \(\hat{\sigma}^{2}(x)\) provides uncertainty information, this uncertainty is not used to trigger individual true objective function evaluation; instead, true evaluations are controlled at the generation level, and the ensemble predictive mean is used to rank candidate solutions during surrogate only CMA-ES generations. The ensemble-based uncertainty quantification (UQ) approach was chosen for its simple integration, parallel training, and efficient warm-start retraining. while being simpler than posterior inference in Bayesian neural networks. Prediction disagreement provides a practical estimate of epistemic uncertainty at moderate computational cost.
\end{enumerate}

\subsection{Adaptive Sampling Strategy}

The AFN uses a model life based decision rule for surrogate-only CMA-ES generations:
\begin{equation}
L_g =
\begin{cases}
4, & \text{if } \tau_g \geq 0.75 \quad (\text{high ranking quality } \rightarrow \text{ more surrogate-only generations}) \\
3, & \text{if } 0.60 \leq \tau_g < 0.75 \\
1, & \text{if } 0.45 \leq \tau_g < 0.60 \\
0, & \text{if } \tau_g < 0.45 \quad (\text{low ranking quality } \rightarrow \text{ true-evaluated generation})
\end{cases}
\end{equation}
where the threshold is based on Kendall's rank correlation $\tau_g$, which compares the previous surrogate predictions with the newly observed true objective values. The thresholds $0.45$, $0.60$, and $0.75$ have been chosen to indicate weak, moderate, and strong ranking agreement, respectively, which control the number of permitted surrogate-only generations.

\section{Experimental Setup}
\subsection{Benchmark Suite }

The proposed AFN-CMA-ES algorithm was evaluated on the COCO/BBOB (Black-Box Optimization Benchmarking) test suite, the standard noiseless benchmark for continuous optimization algorithms. The test suite comprises $24$ noiseless test functions $(f1-f24)$ covering diverse problem characteristics.
\begin{itemize}
\item[]Unimodal separable functions $(f1-f5)$: Sphere, Ellipsoid separable, Rastrigin separable, Büche-Rastrigin, and Linear slope.
\item[]Unimodal low-conditioned functions $(f6-f9)$: Attractive sector, Step-ellipsoid, Rosenbrock original, and Rosenbrock rotated.
\item[]Unimodal high-conditioned functions $(f10-f14)$: Ellipsoid, Discus, Bent cigar, Sharp ridge, and Sum of different powers.
\item[] Multimodal functions $(f15-f24)$: Rastrigin, Weierstrass, Schaffer F7 condition 10, Schaffer F7 condition 1000, Griewank-Rosenbrock, Schwefel, Gallagher 101 peaks, Gallagher 21 peaks, Katsuura, and Lunacek bi-Rastrigin.     
\end{itemize}
\subsection{Test Dimensions}
Experiments were conducted across 4 problem dimensions: $2D$, $3D$, $5D$ and $10D$ to assess the scalability of the proposed method.
\subsection{Compared Algorithms}
AFN-CMA-ES was compared against four state-of-the-art CMA-ES variants:
\begin{enumerate}
    \item CMA-ES: the standard covariance matrix adaptation evolution strategy \cite{hansen2009bipop}.
    \item LQ-CMA-ES: CMA-ES with linear-quadratic surrogate model \cite{hansen2019global}.
    \item DTS-CMA-ES: doubly trained surrogate CMA-ES \cite{pitra2017ordinal}.
    \item LMM-CMA-ES: CMA-ES with local meta-model \cite{auger2013local}.
\end{enumerate}

The algorithms were selected to represent standard, global-surrogate,
ranking-based, and local-surrogate CMA-ES approaches. The baseline results for CMA-ES, LQ-CMA-ES, DTS-CMA-ES, and LMM-CMA-ES were obtained from the publicly available COCO/BBOB data archive. These results were archived and compared with AFN-CMA-ES using the standard COCO/BBOB evaluation protocol, of 15 independent runs of each combination of functions and dimensions.

\subsection{Experimental Configuration}

The evaluation of AFN-CMA-ES was performed on 15 instances in each considered dimension for each function from the COCO/BBOB benchmark set.
For each function dimension target setting, run lengths from the 15 BBOB instances were bootstrapped by resampling runs with replacement. Table~\ref{tab:ert10d} reports dispersion values in parentheses, computed as half the difference between the 90th and 10th percentiles of the bootstrapped run-length distribution. The symbol $\infty$ indicates that the target was not reached within the evaluation budget and the subscript $2500$ denotes the median number of function evaluations conducted when the final target was not reached; in 10D, it corresponds to the budget $250d=2500$. The baseline results for CMA-ES, LQ-CMA-ES, DTS-CMA-ES, and LMM-CMA-ES were obtained from COCO using the same evaluation setting of 15 instances. The maximum evaluation budget was based on COCO/BBOB and scaled with the problem dimension. For AFN-CMA-ES, the surrogate model consisted of 5 MLPs with 2 hidden layers, where the hidden-layer sizes were dimension-dependent. Features and rank-normalized targets were standardized before training. The ANN ensemble was trained using the AdamW optimizer with a learning rate of \(\eta=0.001\) and anchored regularization. The surrogate was retrained using a local archive subset of at most 250 samples selected according to distance from the current CMA-ES mean.

Evaluation budget: The number of evaluations of the true objective function in the evaluation budget \(B = 250d\), where \(d\) is the problem dimension. The factor \(250d\) corresponds to the maximum true-evaluation budget scaled by the problem dimension and includes the initial LHS sampling, true evaluated CMA-ES generations, surrogate retraining and optimization. Therefore, comparisons used the COCO/BBOB noiseless archive ranges: 24 functions \((f_1\text{--}f_{24})\), dimensions 
\(d \in \{2,3,5,10\}\), target precisions \(10^{-8}\), and baselines were compared where corresponding COCO/BBOB archive data were available.

\subsection{Performance Metrics}
The assessment of performance involved COCO standard measures: Runtime distributions: empirical cumulative distribution functions (ECDFs), which are used to describe the fraction of targets that were achieved after a given number of evaluations of the function.

Target precisions: Success rates were reported for different target precisions according to the standard COCO/BBOB post-processing targets up to \(1e{-}8\).

Statistical significance: Statistical comparison can be performed using the COCO/BBOB post processing results to assess whether the observed differences between algorithms were statistically significant.

For true values \(y_i\) and surrogate predictions \(\hat{y}_i\), standard
and rank-normalized RMSE are defined as
\begin{equation}
\begin{aligned}
\mathrm{RMSE}
&=\sqrt{\frac{1}{n}\sum_{i=1}^{n}(y_i-\hat{y}_i)^2},\\
r_i
&=\frac{R_i-1}{N-1},\\
\mathrm{RMSE}_{\mathrm{rank}}
&=\sqrt{\frac{1}{n}\sum_{i=1}^{n}(r_i-\hat{r}_i)^2},
\end{aligned}
\label{eq:rmse}
\end{equation}
where \(R_i\) is the rank of \(y_i\) among \(N\) archive values and
\(\hat{r}_i\) is the prediction on the same normalized-rank scale.
Standard RMSE measures error on the original scale, whereas
rank-normalized RMSE measures dimensionless error on the normalized-rank
scale.

\section{Results and Discussion}

\begin{figure}[ht!]
    \centering

    \includegraphics[width=0.45\textwidth]{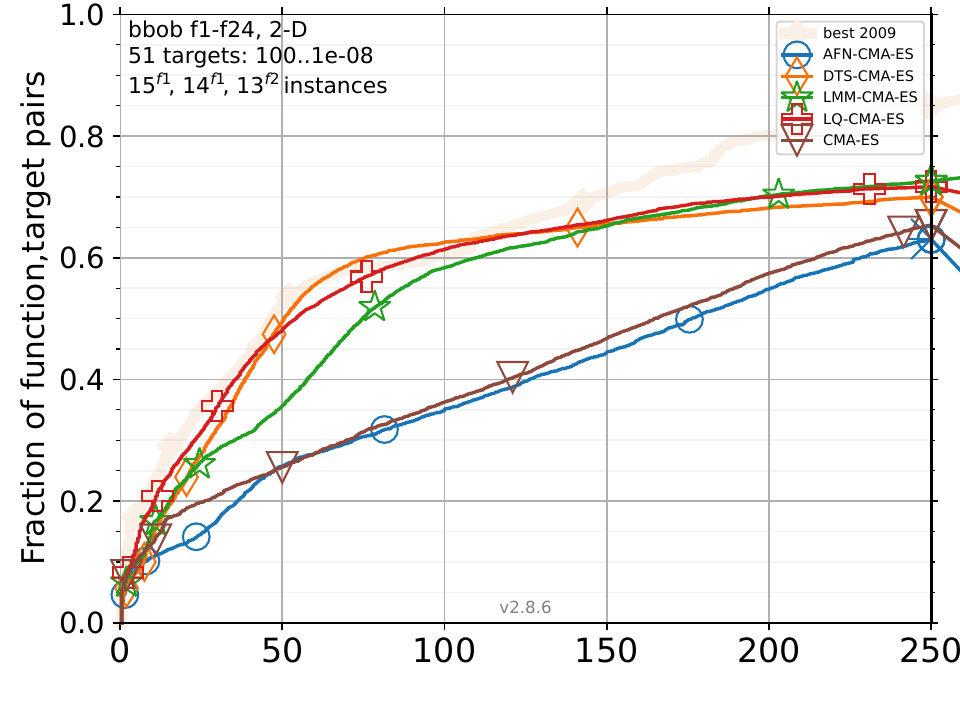}
    \includegraphics[width=0.45\textwidth]{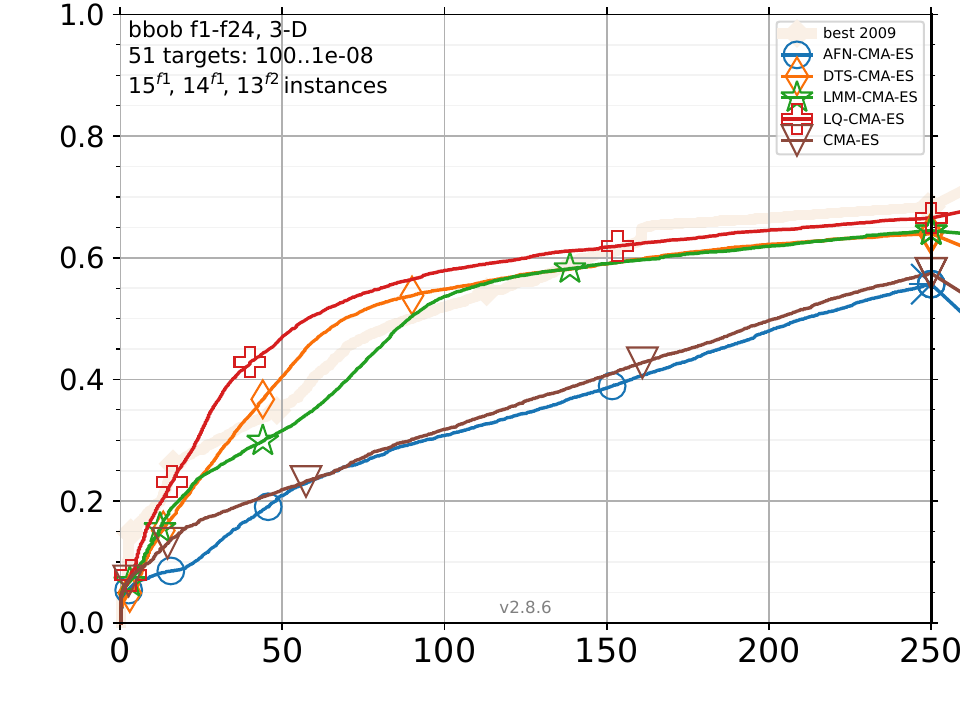}
               
    \includegraphics[width=0.45\textwidth]{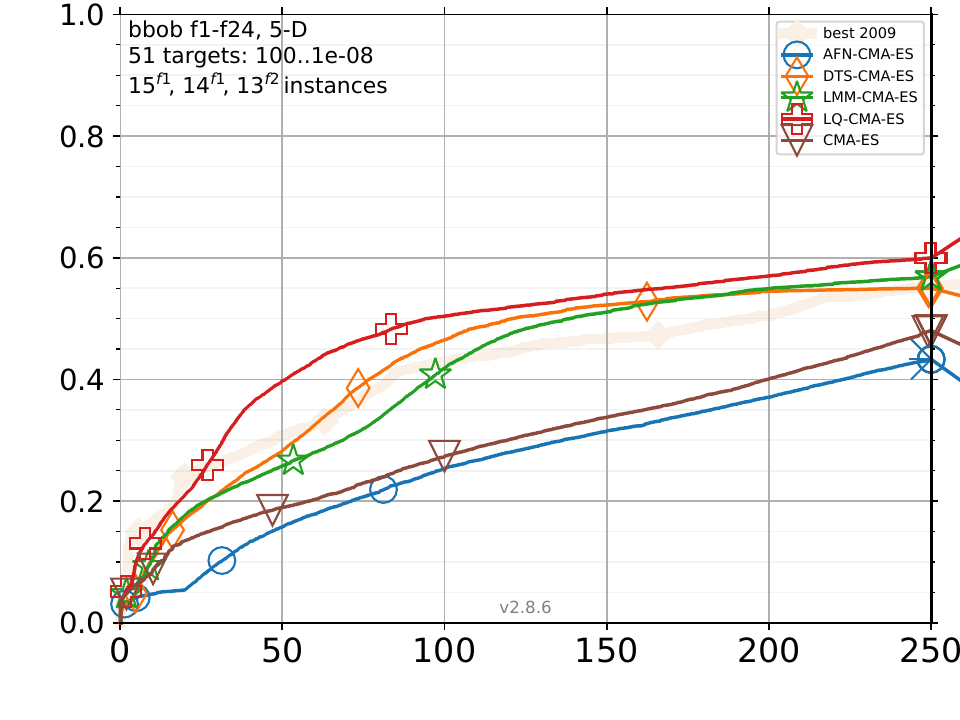}
    \includegraphics[width=0.45\textwidth]{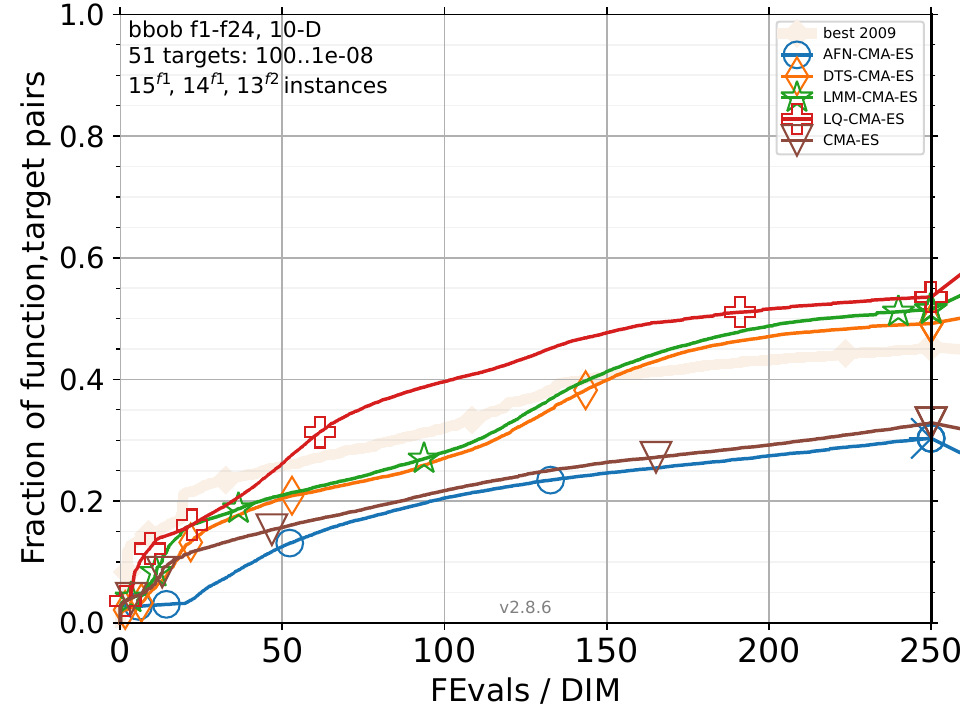}
    
\caption{
Empirical cumulative distribution functions (ECDF) plot analysis of overall performance in four dimensions of the problem. This figure shows ECDFs that have been cumulated over all the 24 noiseless BBOB functions. The subplots concern the following dimensions: (a) $2D$, (b) $3D$, (c) $5D$, and (d) $10D$. The ECDF plots depict all target precision values obtained relative to the number of function evaluations and provide an overall comparison of an algorithm's performance across all dimensions. The results obtained in this case are combined across all functions and targets unless no algorithm among the compared ones achieves the target within the specified evaluation budget. Pairwise Wilcoxon rank-sum tests were used to assess statistical significance between AFN-CMA-ES and the baseline algorithms. For each function-dimension setting, the null hypothesis was that AFN-CMA-ES and the corresponding baseline algorithm had the same distribution of running-time performance values. Statistical significance was assessed at $p<0.05$ after Holm correction
over the 24 BBOB functions.}
    \label{Fig.2.}
\end{figure}

\begin{figure}[ht!]
    \centering

    \includegraphics[width=0.45\textwidth]{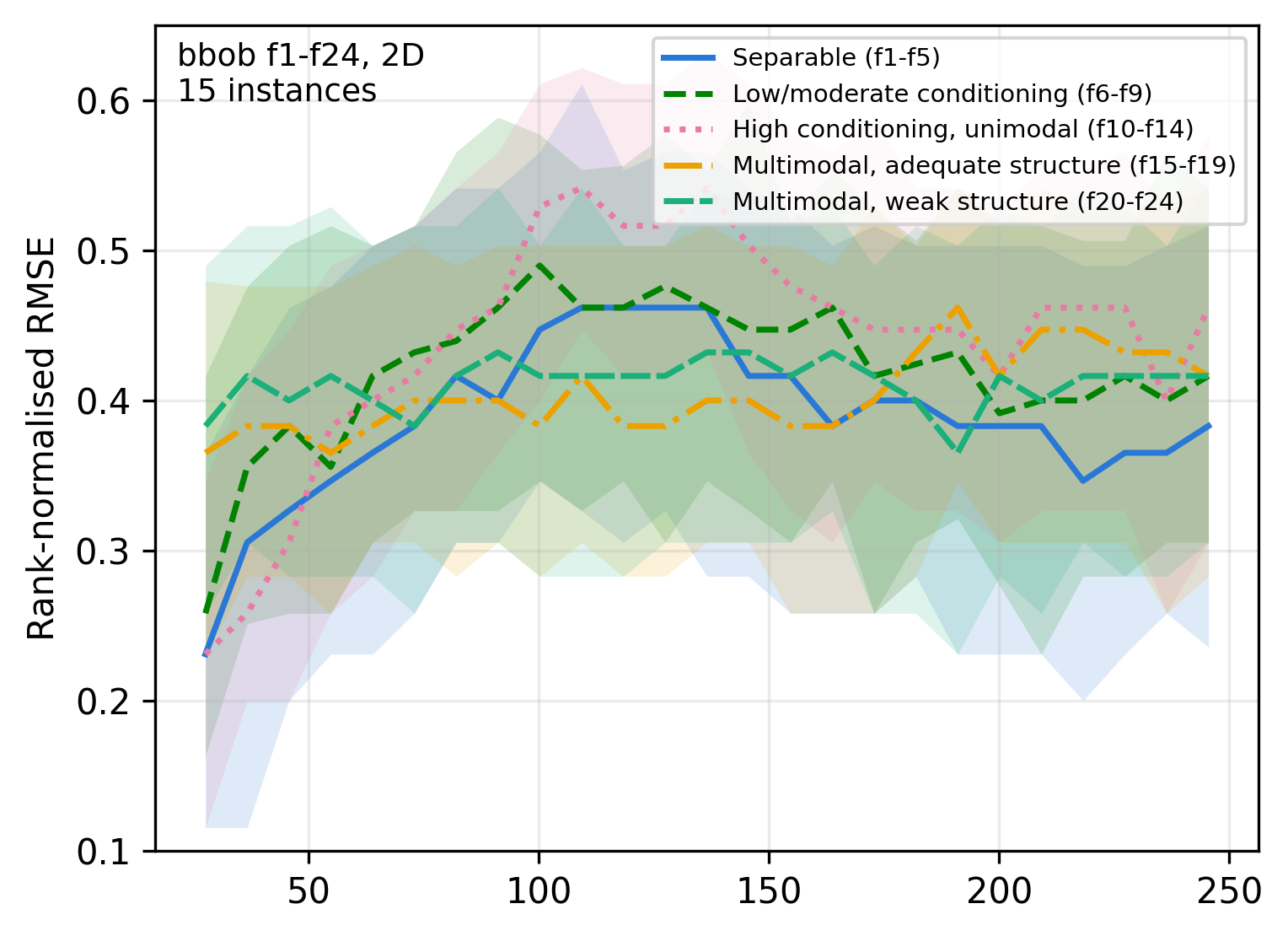}
    \includegraphics[width=0.45\textwidth]{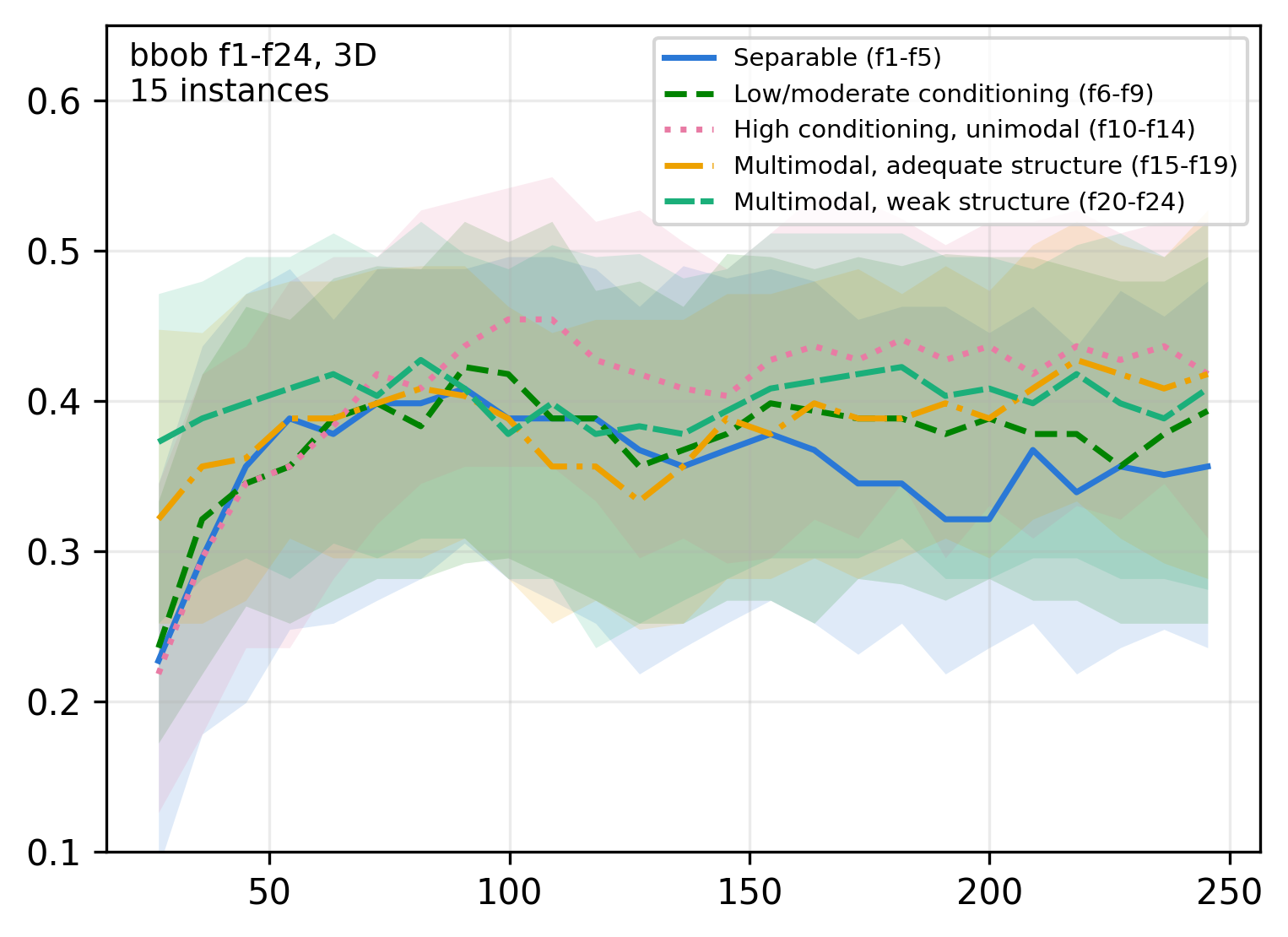}
               
    \includegraphics[width=0.45\textwidth]{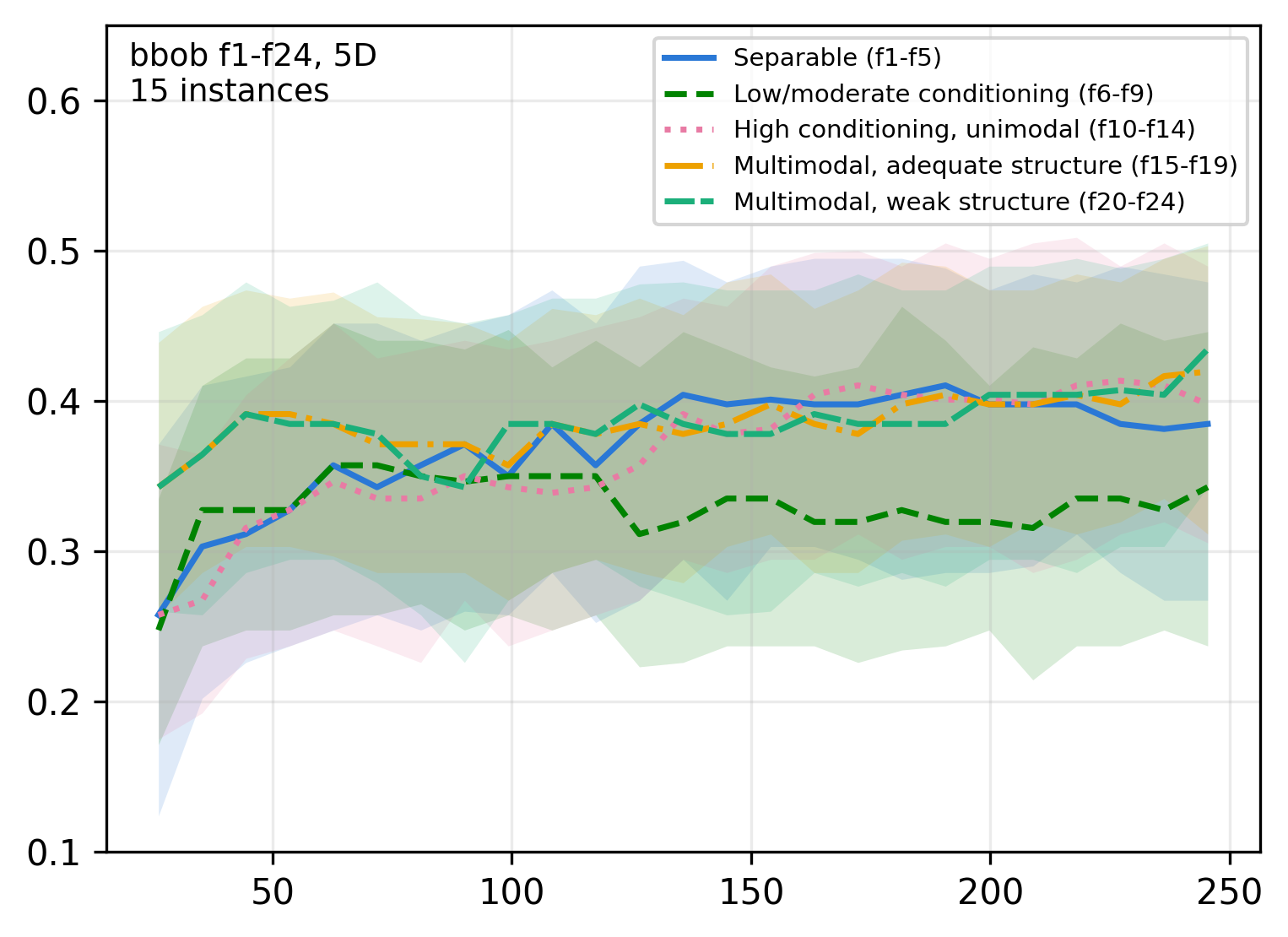}
    \includegraphics[width=0.45\textwidth]{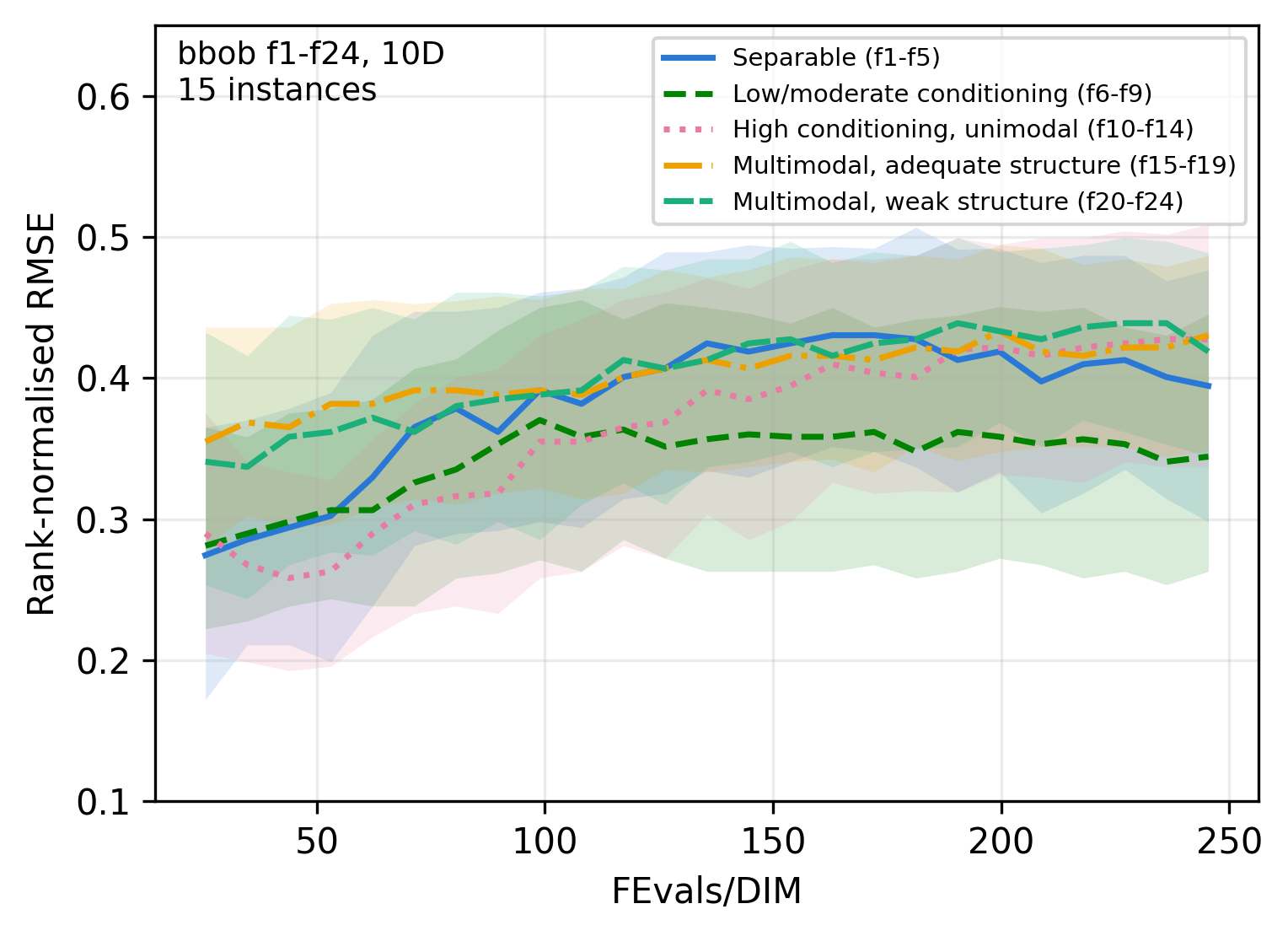}
    
\caption{Evolution of rank-normalized surrogate RMSE with function evaluations per dimension across five BBOB function groups in 2D, 3D, 5D, and 10D. Curves show median RMSE over functions and 15 instances, with shaded interquartile ranges. RMSE generally increases early and then stabilizes, with lower late-stage errors for low and moderately conditioned functions in 5D and 10D.}
 \label{fig:rmse_evolution}
\end{figure}

The overall performance of the compared algorithms across all BBOB functions and target precisions is summarized by the ECDF plots in Fig.~\ref{Fig.2.}. To complement the ECDF analysis, Fig.~\ref{fig:rmse_evolution} shows the evolution of rank-normalized RMSE across five BBOB function groups and four dimensions.

\subsection{Comparison with Baseline Algorithms}

AFN-CMA-ES vs CMA-ES: AFN-CMA-ES achieves performance comparable to the standard CMA-ES baseline on selected unimodal, well-conditioned functions. Surrogate-assisted black-box optimizers are designed to use fewer true objective-function evaluations by replacing part of the evaluation process with surrogate predictions; they are expected to achieve better optimization progress than the original optimizer under the same true-evaluation budget. In this context, comparable performance to CMA-ES can be interpreted as AFN-CMA-ES not behaving in the desired way. In particular, on highly multimodal functions \((f_{15}\text{--}f_{24})\), AFN-CMA-ES performs significantly worse: at 5D, \(f_{15}\) yields 0/15 success for AFN-CMA-ES versus 15/15 for CMA-ES, and at 10D AFN-CMA-ES exhausts its budget \((\infty)\) on \(f_{15}\), \(f_{16}\), \(f_{17}\), and \(f_{18}\).

AFN-CMA-ES vs LQ-CMA-ES: LQ-CMA-ES demonstrates higher performance on most dimensions of separable, low-conditioned functions. On \(f_2\), LQ-CMA-ES achieves expected running time  (ERT) ratios ( algorithm's ERT divided by the COCO/BBOB reference ERT; lower is better, and $\infty$ means failure) of 1.7--2.7 (2D), compared to AFN-CMA-ES's 10--17 with 11/15 success.

AFN-CMA-ES vs DTS-CMA-ES: There is notable variation depending on function type. On Rosenbrock \((f_8)\) across 2D--10D, DTS-CMA-ES performs better than AFN-CMA-ES. AFN-CMA-ES reaches 7/15 success at 2D, 4/15 at 3D, and 0/15 at 5D and 10D.

AFN-CMA-ES vs LMM-CMA-ES: LMM-CMA-ES has a clear advantage on multimodal functions in lower dimensions. On \(f_{15}\) at 2D, LMM-CMA-ES achieves an expected running time ERT ratio of 0.70 with 10 successful trials, while AFN-CMA-ES reaches only 3/15 success.

\subsection{Scalability Analysis}
\begin{figure}[ht!]
    \centering
    \includegraphics[width=0.24\textwidth]{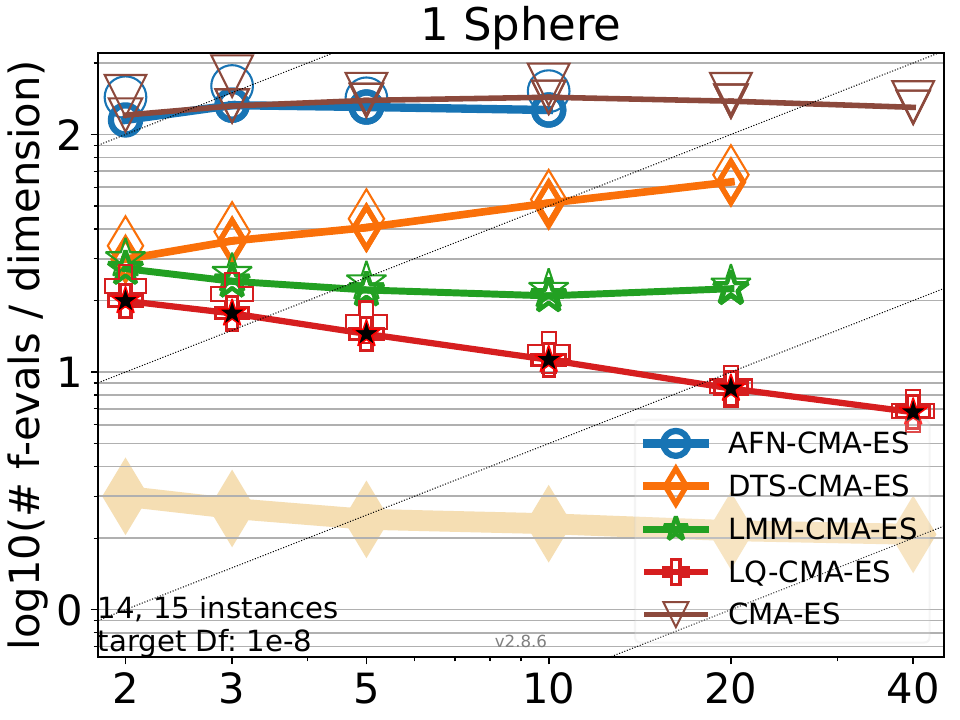}
    \includegraphics[width=0.24\textwidth]{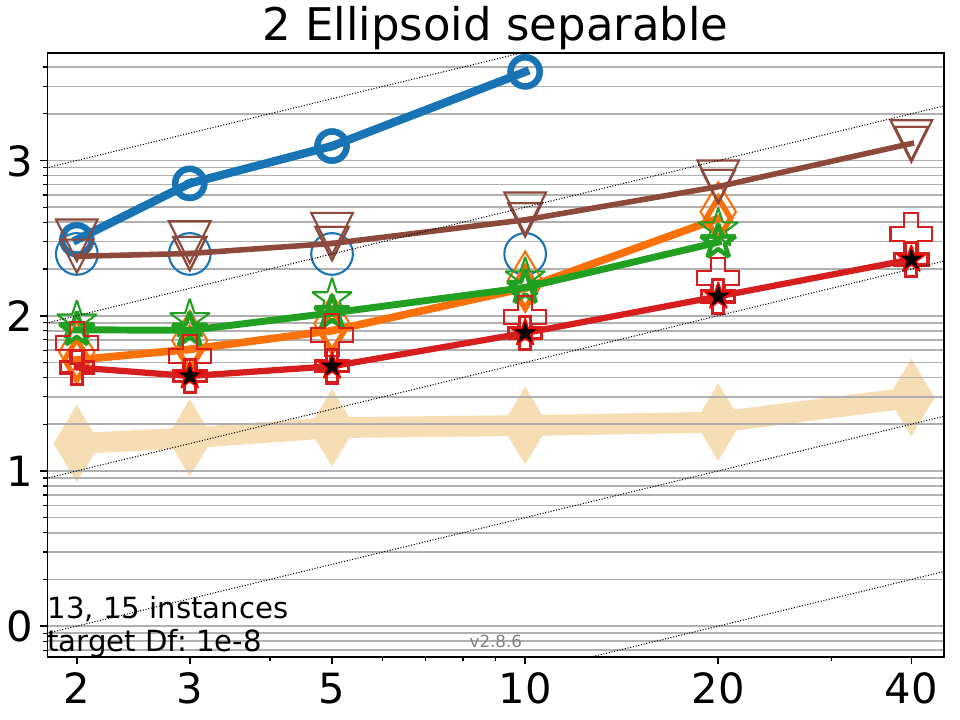}
    \includegraphics[width=0.24\textwidth]{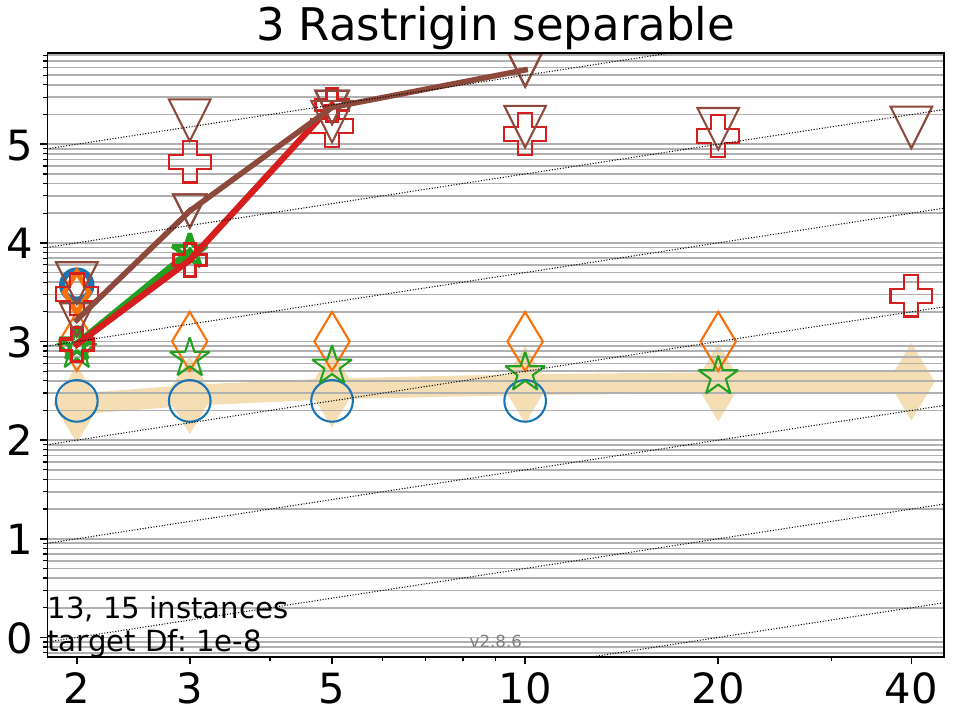}
    \includegraphics[width=0.24\textwidth]{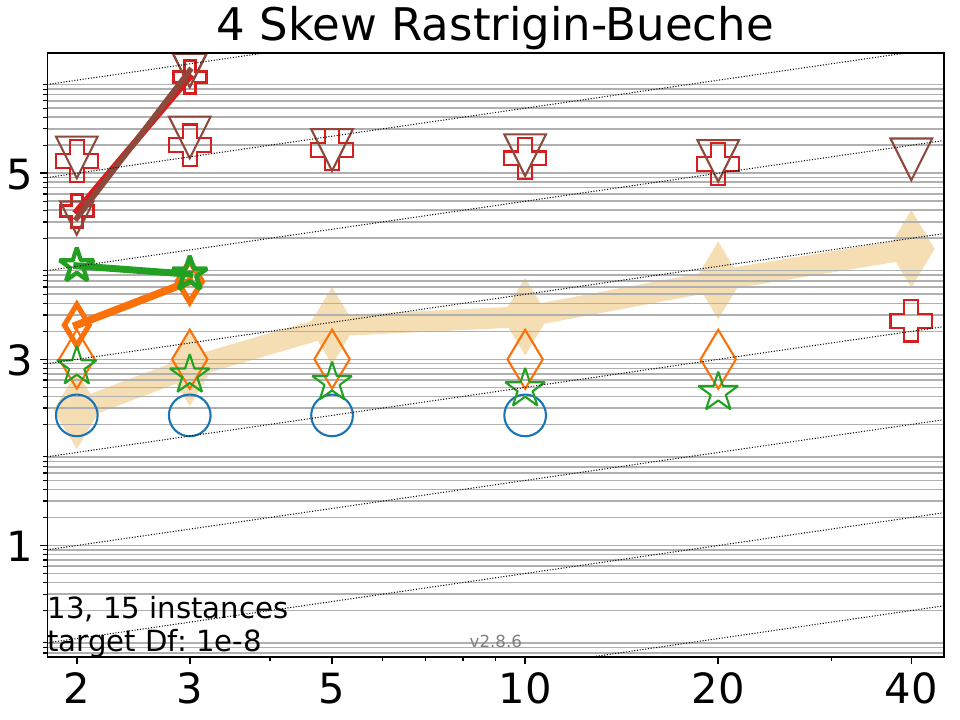}

    \includegraphics[width=0.24\textwidth]{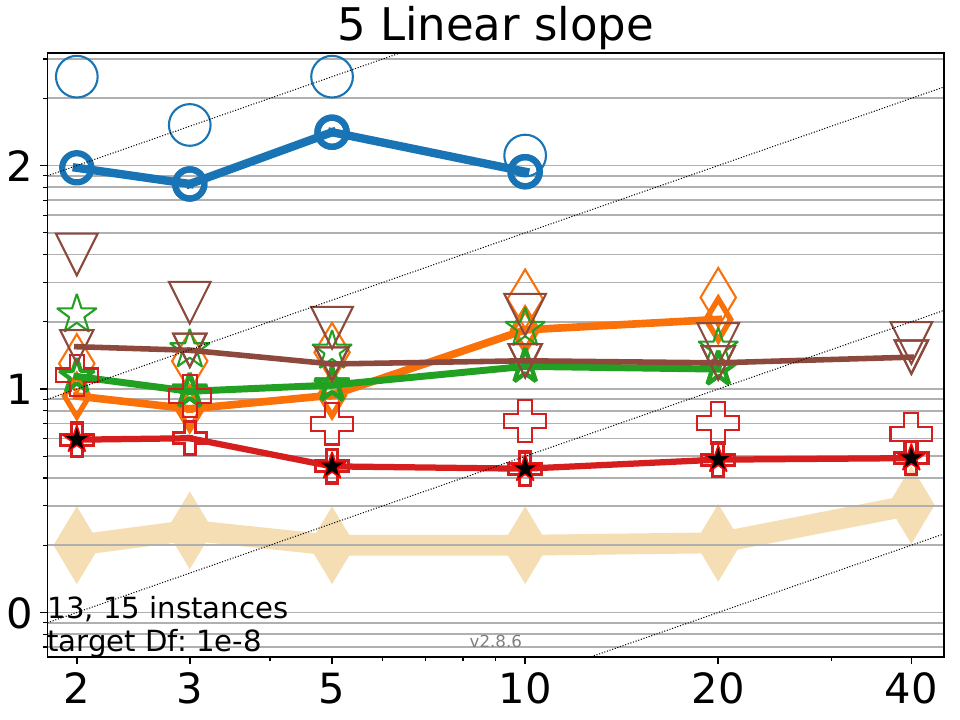}
    \includegraphics[width=0.24\textwidth]{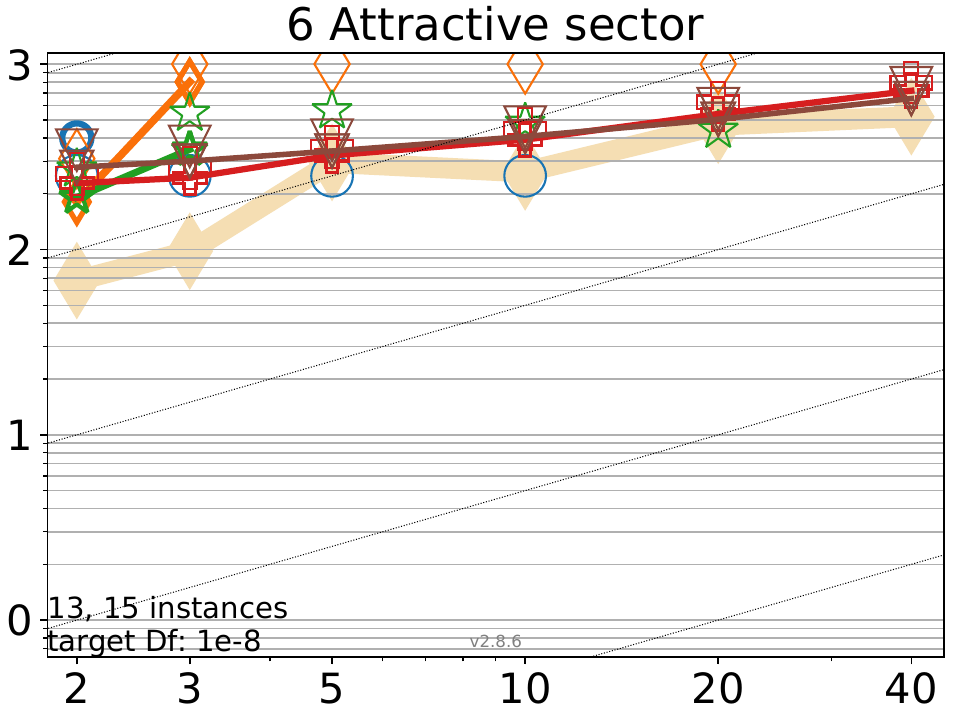}
    \includegraphics[width=0.24\textwidth]{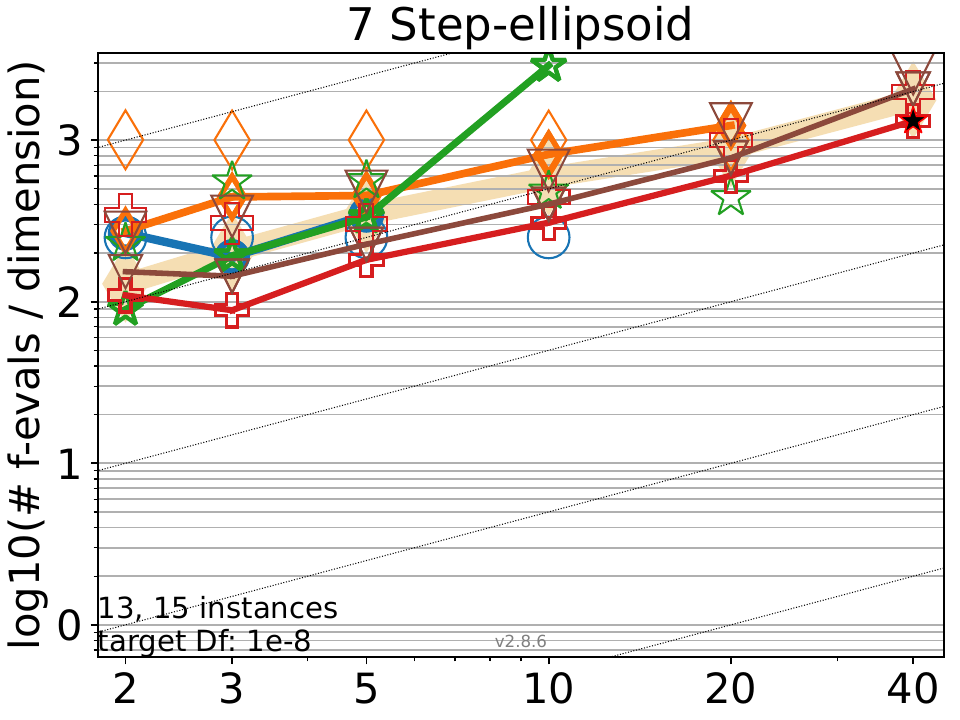}
    \includegraphics[width=0.24\textwidth]{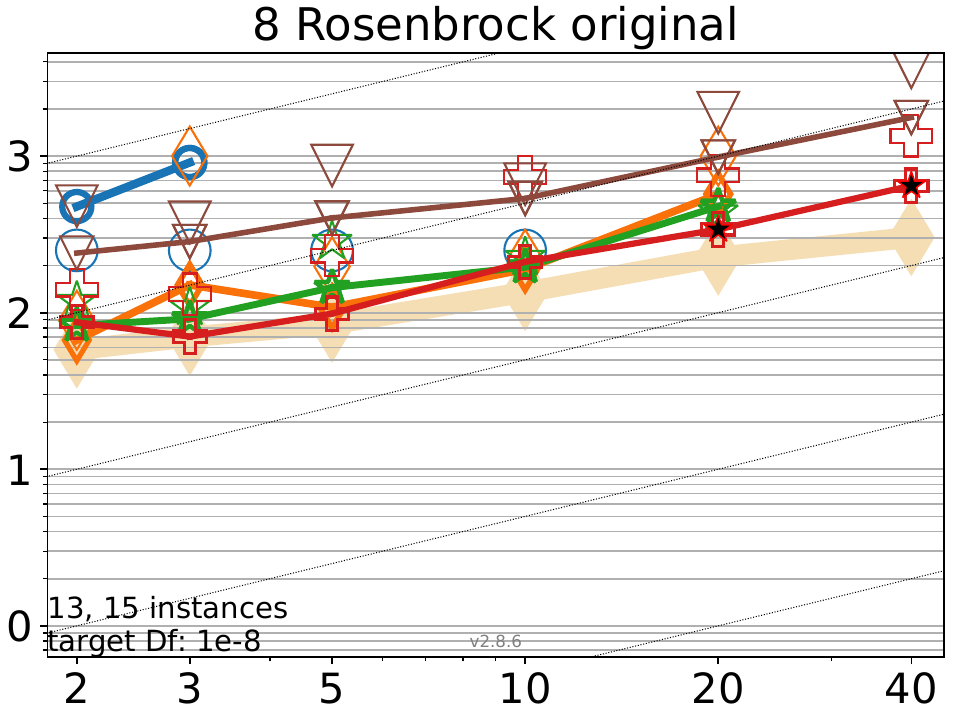}

    \includegraphics[width=0.24\textwidth]{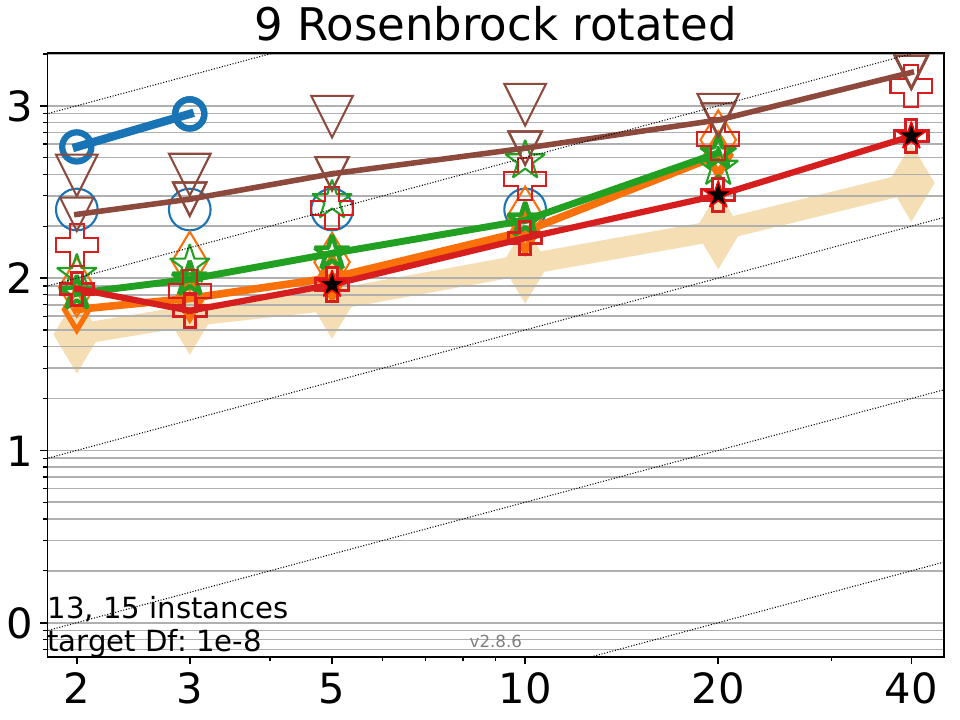}
    \includegraphics[width=0.24\textwidth]{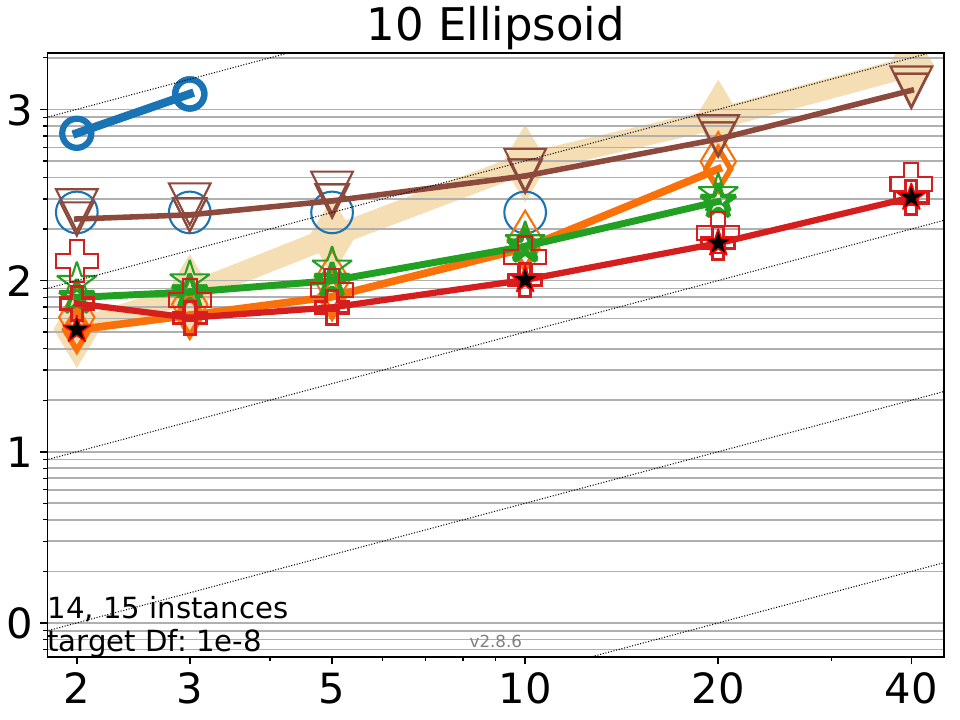}
    \includegraphics[width=0.24\textwidth]{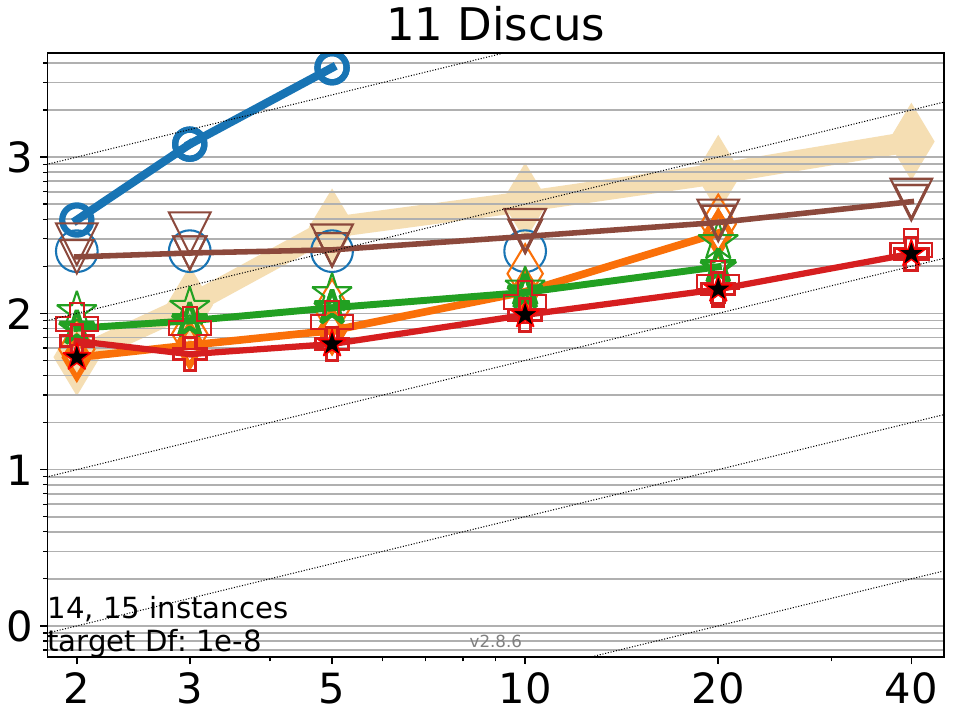}
    \includegraphics[width=0.24\textwidth]{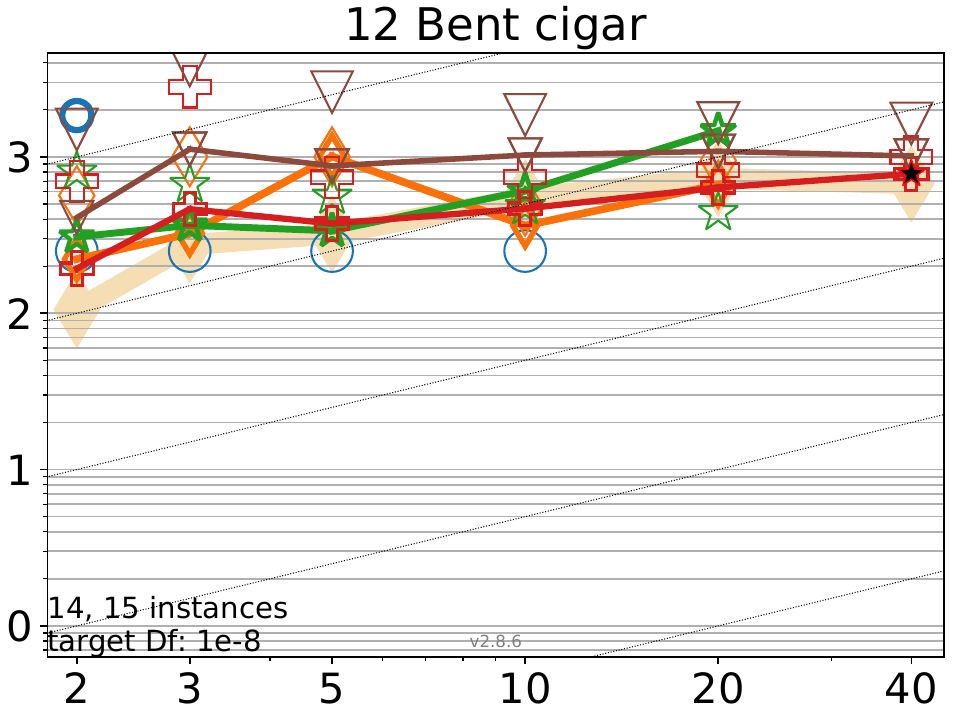}

    \includegraphics[width=0.24\textwidth]{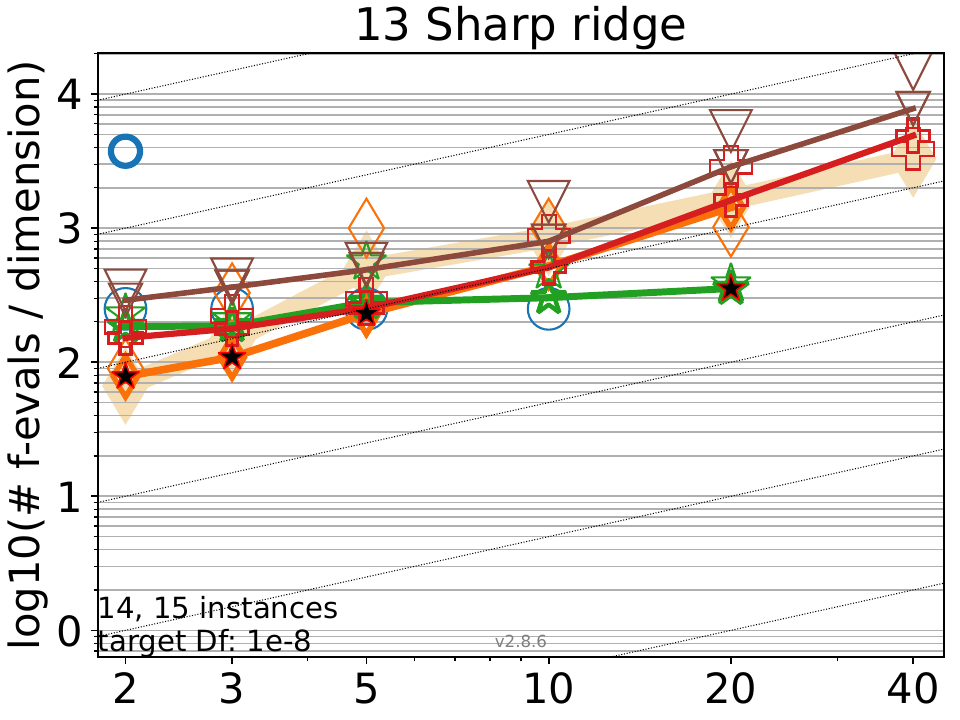}
    \includegraphics[width=0.24\textwidth]{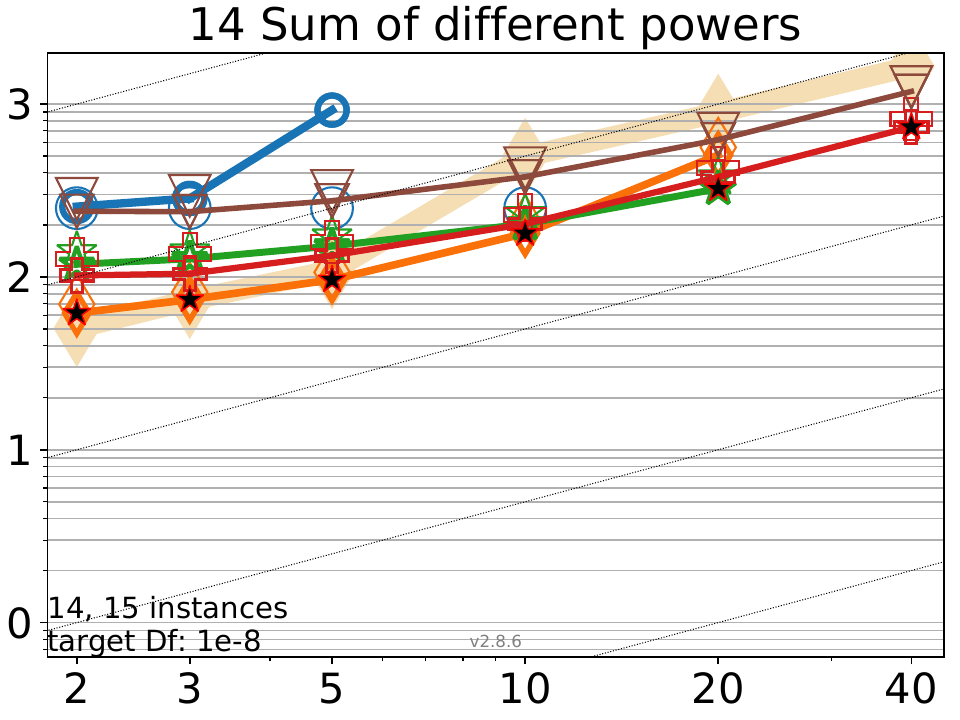}
    \includegraphics[width=0.24\textwidth]{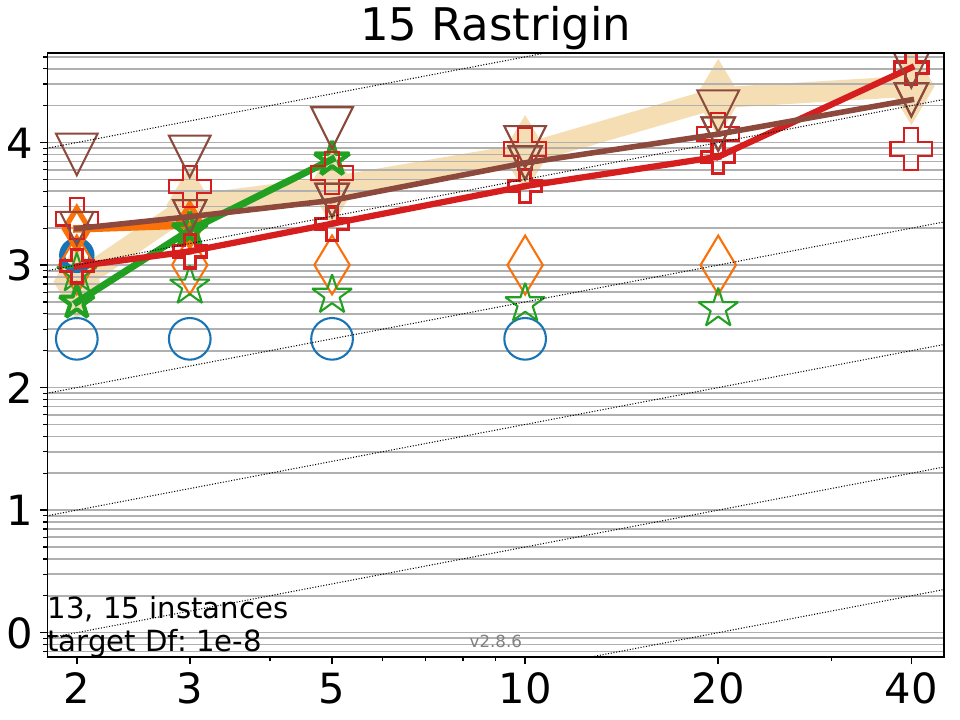}
    \includegraphics[width=0.24\textwidth]{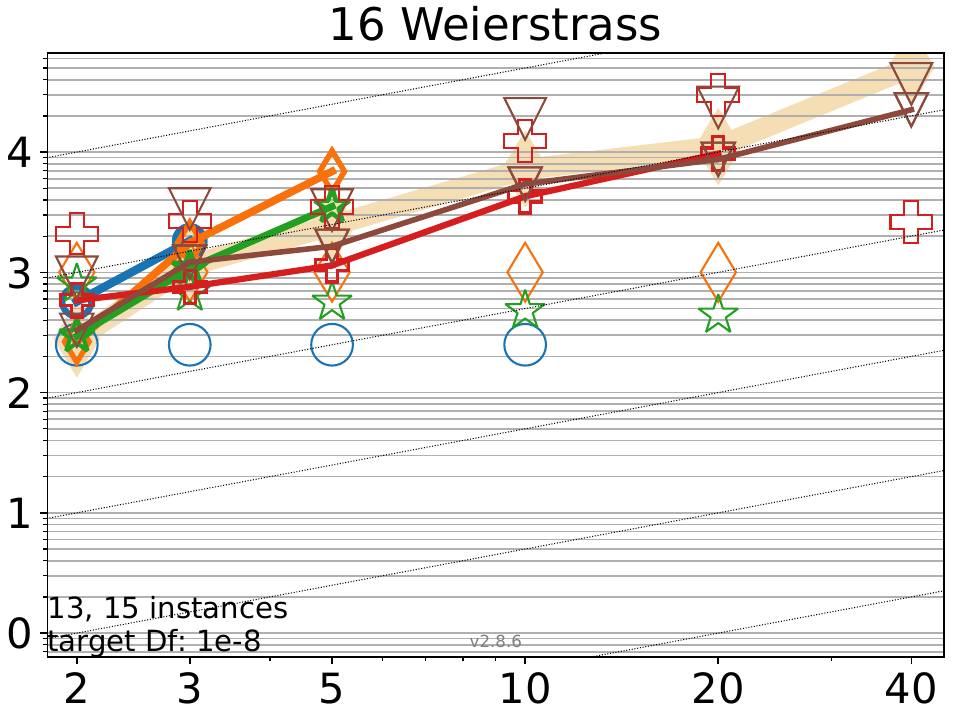}

    \includegraphics[width=0.24\textwidth]{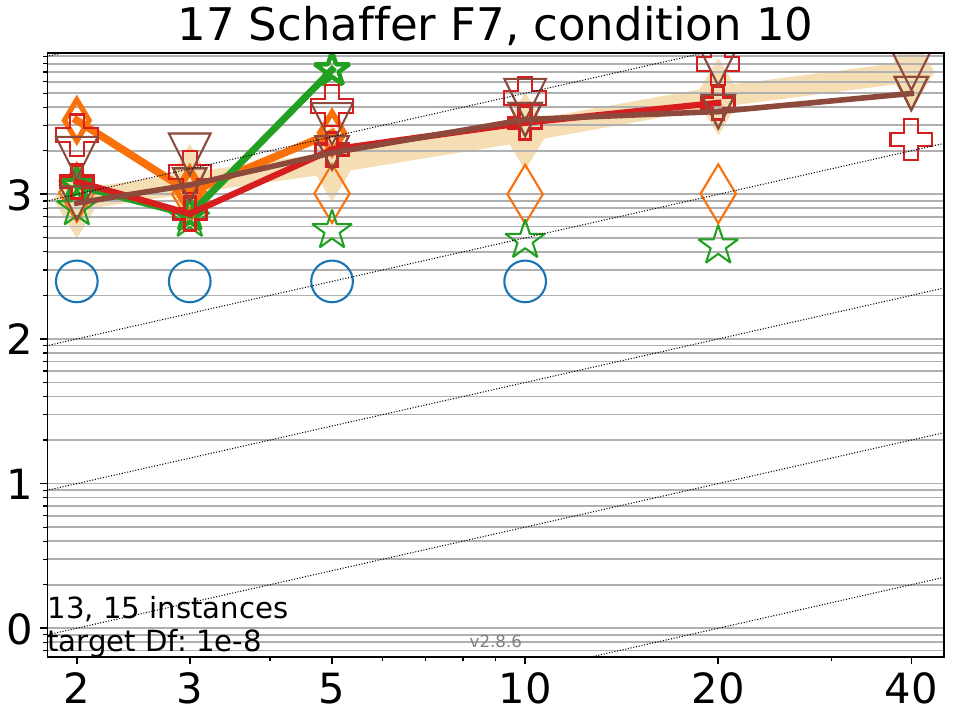}
    \includegraphics[width=0.24\textwidth]{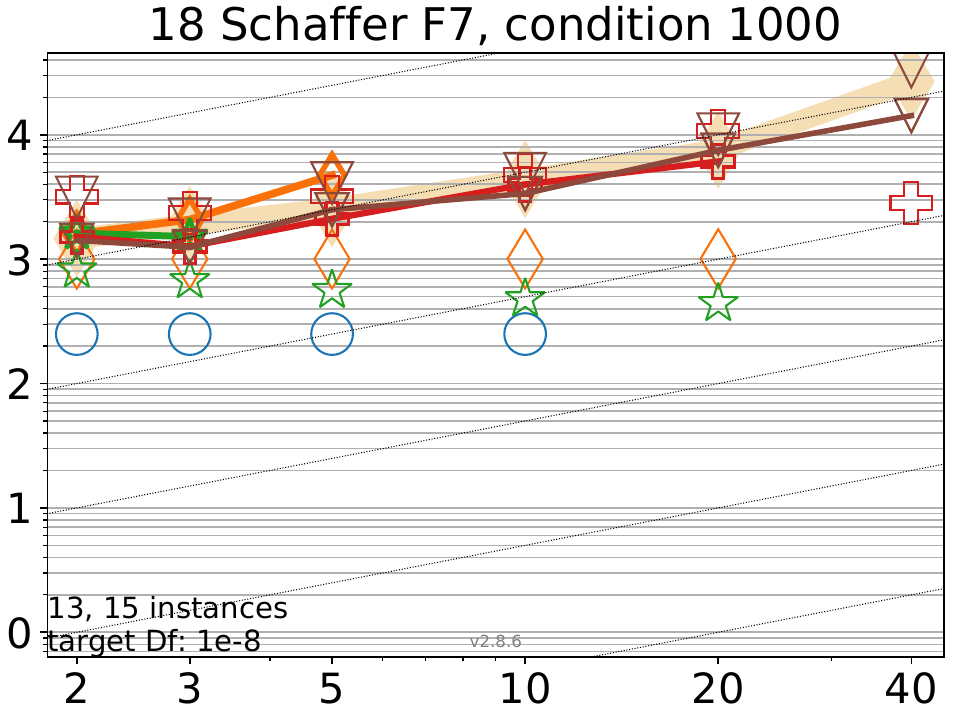}
    \includegraphics[width=0.24\textwidth]{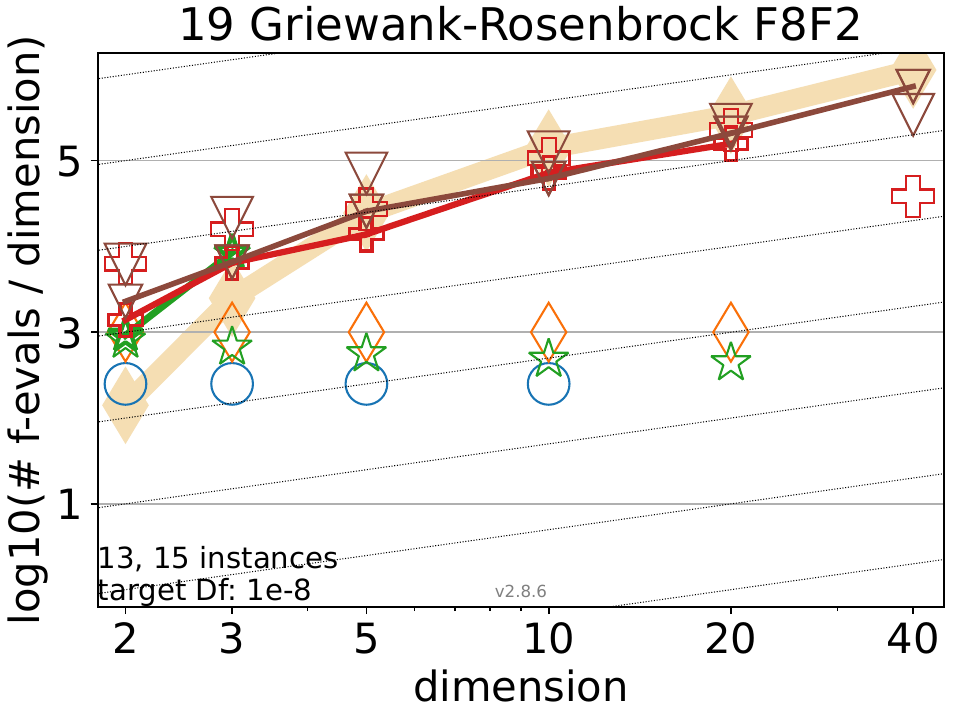}
    \includegraphics[width=0.24\textwidth]{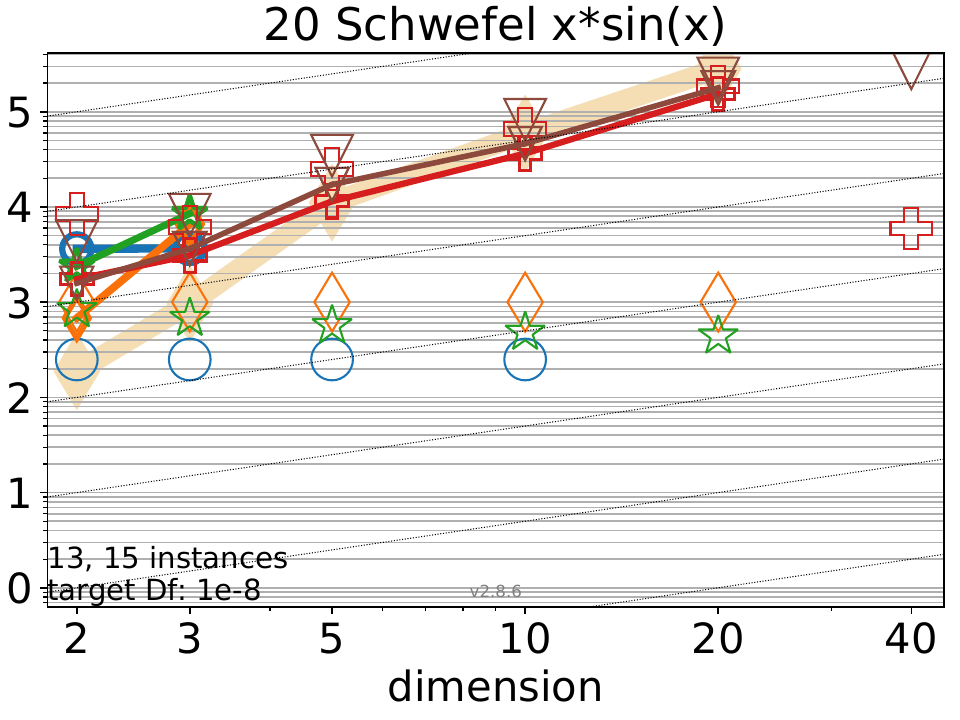}

    \includegraphics[width=0.24\textwidth]{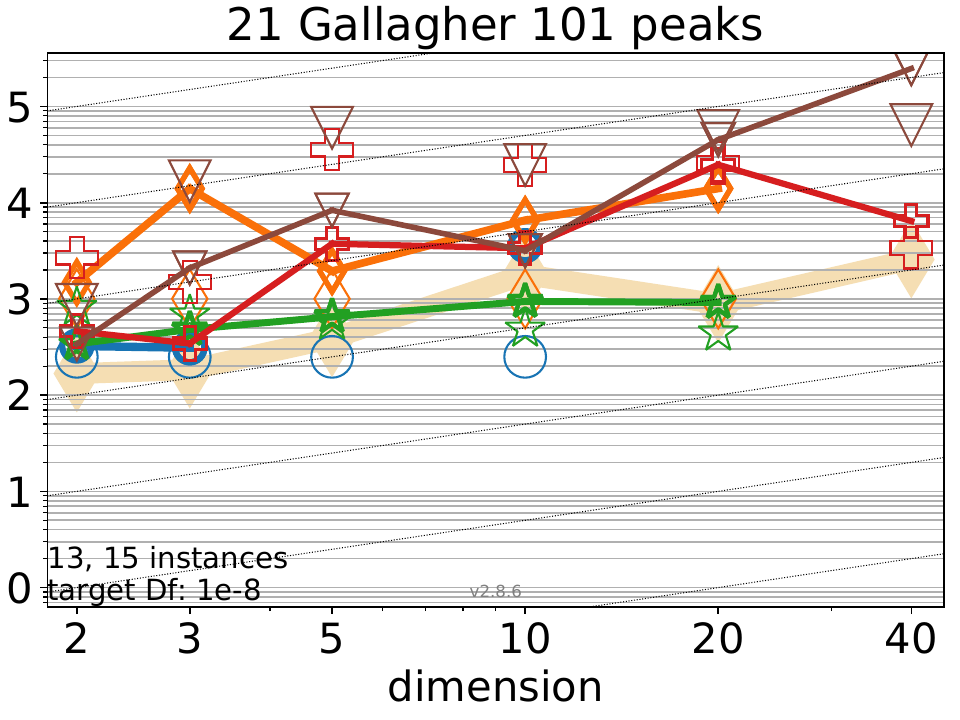}
    \includegraphics[width=0.24\textwidth]{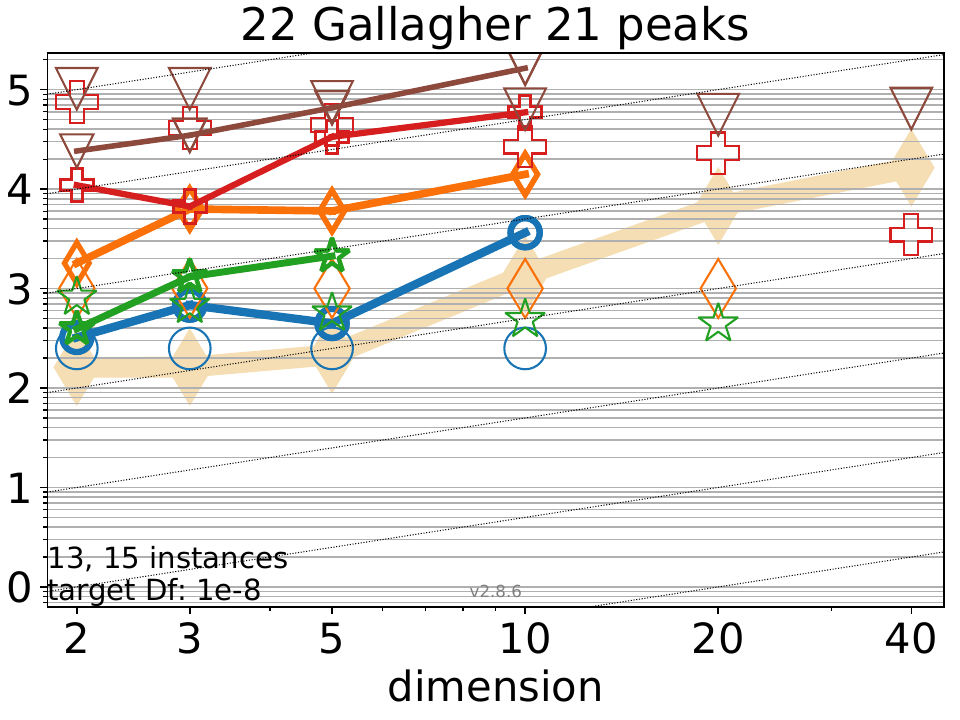}
    \includegraphics[width=0.24\textwidth]{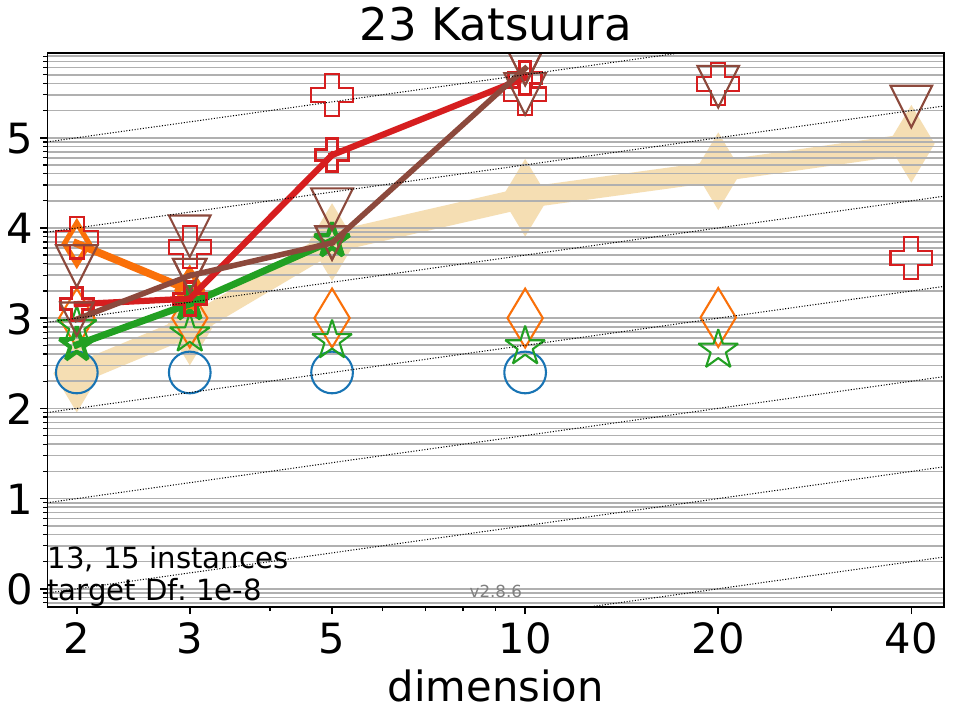}
    \includegraphics[width=0.24\textwidth]{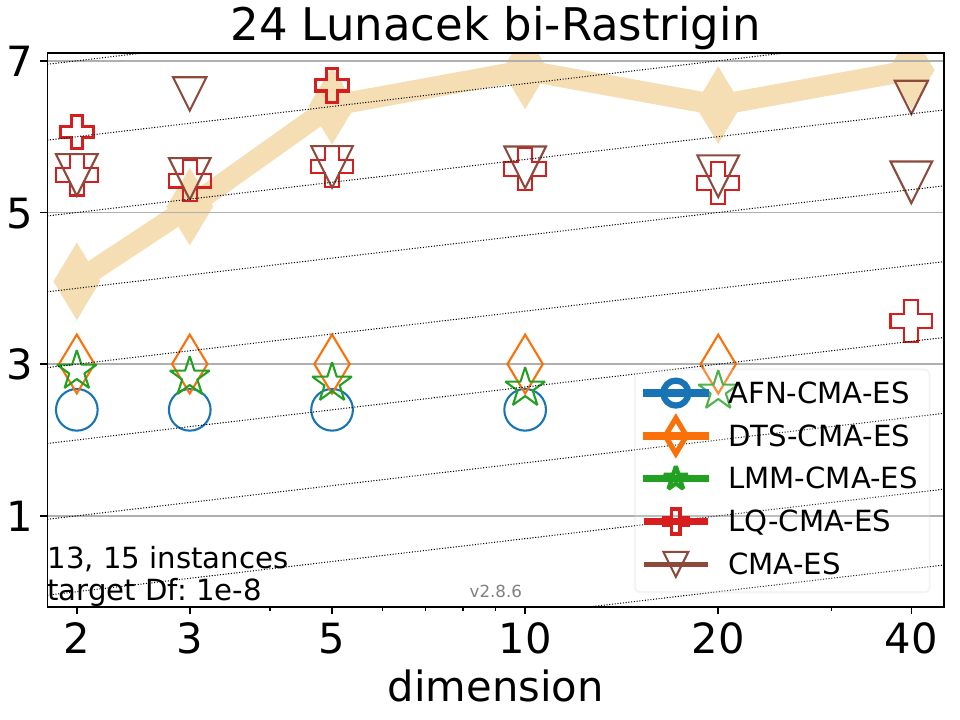}
    \caption{ ERT in number of $f$-evaluations as $\log_{10}$ of the value divided by dimension for target function value $10^{-8}$ versus dimension. Slanted grid lines indicate quadratic scaling with the dimension. Different symbols correspond to different algorithms given in the legend of $f1$ and $f24$. Light symbols give the maximum number of evaluations from the longest trial divided by dimension. 
    Legend: {\color{blue}$\circ$}: AFN-CMA-ES, 
    {\color{Orange}$\diamond$}: DTS-CMA-ES, 
    {\color{green}$\star$}: LMM-CMA-ES, 
    {\color{red}$+$}: LQ-CMA-ES, 
    {\color{Brown}$\nabla$}: CMA-ES}
    \label{Fig. 1.}
\end{figure}

Scalability analysis in this case considers expected running time ERT ratios aggregated across all 24 BBOB functions and examines the dimensional scaling behaviour. The dimensional scaling behaviour of the compared algorithms is shown in Fig.~\ref{Fig. 1.}. On the Sphere function \((f_1)\), AFN-CMA-ES expected running time ERT ratios increase from \(34\) in 2D to \(49\) in 10D, with 15/15 success across the available dimensions. In contrast, LQ-CMA-ES ratios remain relatively stable from \(4.5\) to \(6.2\) across 2D to 10D, indicating that its linear quadratic surrogate scales efficiently on separable functions, where the additive problem structure aligns well with the surrogate's inductive bias.

The scaling gap becomes larger on non-separable functions \((f_8, f_9)\), which are well-conditioned. The value of AFN-CMA-ES at 10D is \(\infty\) \((f_8)\) and \(\infty\) \((f_9)\) versus \(1.4\) and \(1.5\) for DTS-CMA-ES, which represents a larger relative difference than on separable functions.

The most significant scalability failure is on highly multimodal functions \((f_{15}\text{--}f_{24})\) in higher dimensions. At 10D, AFN-CMA-ES exhausts its budget \((\infty)\) on \(f_{15}\), \(f_{16}\), \(f_{17}\), \(f_{18}\), \(f_{19}\), \(f_{20}\), \(f_{23}\), and \(f_{24}\). Predictions were compared with true objective values before retraining. The root mean squared error (RMSE) measured prediction error, while Kendall's $\tau$ measured ranking accuracy. AFN-CMA-ES may perform poorly on multimodal and higher-dimensional problems because the limited local archive becomes sparse, the MLP may not capture complex landscapes, and overlapping bootstrap samples may reduce ensemble reliability.

\subsection{Statistical Analysis}

Wilcoxon rank-sum tests ($p < 0.05$, Holm corrected for 24 functions) confirm LQ-CMA-ES makes statistically significant improvements over all other available algorithms on separable functions across most dimensions. LQ-CMA-ES achieves high significance levels  on \(f_1\), \(f_2\), \(f_5\), and \(f_{10}\) across the available dimensions where corresponding archive data are available, demonstrating strong superiority on these function classes. AFN-CMA-ES does not achieve statistically significant improvements over any baseline algorithm on any function dimension combination in the current results. On some individual functions, AFN-CMA-ES shows comparable performance to selected surrogate-assisted baselines. AFN-CMA-ES performs worst relative to the baselines on multimodal functions \(f_{15}\)--\(f_{24}\) at dimensions \(\geq 5D\). Since these failures reduce the number of target values reached within the evaluation budget, they lower the overall aggregate ECDF performance. This implies that the global MLP ensemble surrogate, although based on ranking-based generation control, lacks the selectivity benefit to offset its computational cost in intricate landscapes. Further studies on selecting between local and global surrogate models based on observed landscape structure should be conducted in the future.

\begin{table}[ht!]
\centering
\scriptsize
\setlength{\tabcolsep}{3pt}
\renewcommand{\arraystretch}{0.88}
\caption{Expected runtime (ERT) ratios in 10-D, measured as the number of function evaluations divided by the reference ERT, where reference ERT denotes the best BBOB-2009 ERT from the COCO/BBOB archive for the same function, dimension, and target precision. Values in parentheses denote the dispersion measure, computed as half difference between 10\% and 90\% of bootstrapped run lengths. The column \#succ reports the number of trials reaching \(f_{\mathrm{opt}} + 10^{-8}\).}
\label{tab:bbob_ert_2d}
\label{tab:ert10d}
\begin{tabular}{lcccccccc}
\toprule
Function & $1\text{e}1$ & $1\text{e}0$ & $1\text{e}{-1}$ & $1\text{e}{-2}$ & $1\text{e}{-3}$ & $1\text{e}{-5}$ & $1\text{e}{-7}$ & \#succ \\
\midrule

\multicolumn{9}{l}{\textbf{10-D}} \\
\midrule

f1 ref. ERT & 22 & 23 & 23 & 23 & 23 & 23 & 23 & 15/15 \\
AFN-CMA-ES & 10 (0.8) & 14 (1) & 18 (2) & 22 (1) & 27 (3) & 38 (5) & 49 (8) & 15/15 \\

f2 ref. ERT & 187 & 190 & 191 & 191 & 193 & 194 & 195 & 15/15 \\
AFN-CMA-ES & 13 (9) & 18 (12) & 93 (78) & 94 (82) & 93 (97) & 94 (93) & 94 (103) & 1/15 \\

f3 ref. ERT & 1739 & 3600 & 3609 & 3636 & 3642 & 3646 & 3651 & 15/15 \\
AFN-CMA-ES &$\infty$  & $\infty$  & $\infty$  & $\infty$  & $\infty$  & $\infty$  & $\infty2500$ & 0/15 \\

f4 ref. ERT & 2234 & 3626 & 3660 & 3695 & 3707 & 3744 & 28767 & 12/15 \\
AFN-CMA-ES & $\infty$ & $\infty$ & $\infty$ & $\infty$ & $\infty$ & $\infty$ & $\infty2500$  & 0/15 \\

f5 ref. ERT & 20 & 20 & 20 & 20 & 20 & 20 & 20 & 15/15 \\
AFN-CMA-ES & 14 (1.0) & 17 (0.9) & 19 (1) & 21 (2) & 23 (3) & 28 (4) & 38 (4) & 15/15 \\

f6 ref. ERT & 412 & 623 & 826 & 1039 & 1292 & 1841 & 2370 & 15/15 \\
AFN-CMA-ES & 1.6 (0.3) & 1.7(0.2) & 1.7(0.3) & 1.7 (0.3) & 1.8 (0.3) & 20 (25) & $\infty2500$ & 0/15 \\

f7 ref. ERT & 172 & 1611 & 4195 & 5099 & 5141 & 5141 & 5389 & 15/15 \\
AFN-CMA-ES & 2.2 (0.5) &1.7 (2) & 2.9 (3) & 7.3 (7) & 7.2 (7) & 7.2 (6) & $\infty2500$ & 0/15 \\

f8 ref. ERT & 326 & 921 & 1114 & 1217 & 1267 & 1315 & 1343 & 15/15 \\
AFN-CMA-ES & 2.1 (0.8) & 20 (30)& $\infty$ & $\infty$ & $\infty$ & $\infty$ & $\infty2500$ & 0/15 \\

f9 ref. ERT & 200 & 648 & 857 & 993 & 1065 & 1138 & 1185 & 15/15 \\
AFN-CMA-ES & 3.5(2) & 6.4(5) & 14 (17)& 19 (17) & 18 (14) & $\infty$ & $\infty2500$ & 0/15 \\

f10 ref. ERT & 1835 & 2172 & 2455 & 2728 & 2802 & 4543 & 4739 & 15/15 \\
AFN-CMA-ES & 3.9(4) & $\infty$ & $\infty$ & $\infty$ & $\infty$  & $\infty$ & $\infty2500$ & 0/15 \\

f11 ref. ERT & 266 & 1041 & 2602 & 2954 & 3338 & 4092 & 4843 & 15/15 \\
AFN-CMA-ES & 13 (10) & 6.9 (6) & 4.7 (5) & 6.3 (5) & 11 (12) & $\infty$ & $\infty2500$ & 0/15 \\

f12 ref. ERT & 515 & 896 & 1240 & 1390 & 1569 & 3660 & 5154 & 15/15 \\
AFN-CMA-ES & 2.2 (1) & 12 (17) & 15 (25) & $\infty$ & $\infty$ & $\infty$ & $\infty2500$ & 0/15 \\

f13 ref. ERT & 387 & 596 & 797 & 1014 & 4587 & 6208 & 7779 & 15/15 \\
AFN-CMA-ES & 1.7 (0.4) & 5.3 (4) & 7.0 (4) & 12 (16) & $\infty$ & $\infty$ & $\infty2500$ & 0/15 \\

f14 ref. ERT & 37 & 98 & 133 & 205 & 392 & 687 & 4305 & 15/15 \\
AFN-CMA-ES & 5.4(2) & 3.6 (0.5) & 3.7 (0.4) & 3.5 (0.9) & 2.9 (0.7) & 3.0 (0.4) & $\infty2500$ & 0/15 \\

f15 ref. ERT & 4774 & 39246 &  73643 & 74669 & 75790  & 77814 & 79834  & 12/15 \\
AFN-CMA-ES & 3.7 (4) & $\infty$ & $\infty$ & $\infty$ & $\infty$ & $\infty$ & $\infty2500$ & 0/15 \\

f16 ref. ERT & 425 & 7029 & 15779 & 45669 & 51151 & 65798 & 71570 & 15/15 \\
AFN-CMA-ES & 2.2 (0.7) & 0.25(0.2) & 0.54(0.4) & 0.81 (0.8) & $\infty$& $\infty$ &$\infty2500$ & 0/15 \\

f17 ref. ERT & 26 & 429 & 2203 & 6329 & 9851 & 20190 & 26503 & 15/15 \\
AFN-CMA-ES & 4.7 (4) &1.4(0.4)& \(0.57\,(0.1)\uparrow 1\) & 0.52(0.4) & 0.89(1.0) & $\infty$ & $\infty2500$ & 0/15 \\

f18 ref. ERT & 238 & 836 & 7012 & 15928 & 27536 & 37234 & 42708 & 15/15 \\
AFN-CMA-ES & 1.5 (0.5) & 1.8 (1) & 5.3 (6) & $\infty$ & $\infty$ & $\infty$ & $\infty2500$ & 0/15 \\

f19 ref. ERT & 1 & 1 & 10609 & 9.8e5 & 1.4e6 & 1.4e6 & 1.4e6 & 15/15 \\
AFN-CMA-ES & 216 (67) &  $\infty$ &  $\infty$ &  $\infty$&  $\infty$ &  $\infty$ & $\infty2500$& 0/15 \\

f20 ref. ERT & 32 & 15426 & 5.5e5 & 5.7e5 & 5.7e5 & 5.8e5 & 5.9e5 & 15/15 \\
AFN-CMA-ES & 7.9 (0.6) & $\infty$ & $\infty$ & $\infty$ & $\infty$ & $\infty$ & $\infty2500$ & 0/15 \\

f21 ref. ERT & 130 & 2236 & 4392 & 4487 & 4618 & 5074 & 11329 & 15/15 \\
AFN-CMA-ES & 7.2 (10) & 4.7 (3) & 8.1 (14) & 7.9 (8) & 7.7 (9) & 7.0 (9) & 3.1 (3) & 1/15 \\

f22 ref. ERT & 98 & 2839 & 6353 & 6620 & 6798 & 8296 & 10351 & 15/15 \\
AFN-CMA-ES & 7.3 (13) & 1.9 (3) & 5.6 (6) & 5.4 (5) & 5.3 (7) & 4.4 (6) & 3.5 (3) & 1/15 \\

f23 ref. ERT & 3.0 & 915 & 16425 & 1.8e5 & 2.0e5 & 2.1e5 & 2.1e5 & 15/15 \\
AFN-CMA-ES & 1.6 (2)  & $\infty$  & $\infty$  & $\infty$  & $\infty$  & $\infty$  &  $\infty2500$  & 0/15 \\

f24 ref. ERT & 98761 & 1.0e6 & 7.5e7 & 7.5e7 & 7.5e7 & 7.5e7 & 7.5e7 & 1/15 \\
AFN-CMA-ES & $\infty$  & $\infty$  & $\infty$  & \(\infty\) & \(\infty\) & \(\infty\) & $\infty2500$ & 0/15 \\

\bottomrule
\end{tabular}
\end{table}

\section{Conclusion}
Surrogate assisted black-box optimization addresses the small data challenge that arises when each objective function evaluation is computationally costly, limiting the total number of evaluations available. In this setting, the surrogate model must approximate the objective function from a restricted archive of evaluated samples, making data efficiency essential. The experimental assessment of the proposed AFN-CMA-ES on the COCO/BBOB benchmark suite shows that the proposed ANN based surrogate assisted framework can reduce the reliance on direct objective function evaluations in selected problem classes, but its performance remains strongly dependent on the objective function landscape structure. The results indicate that AFN-CMA-ES achieves comparable performance to selected surrogate assisted CMA-ES variants on some unimodal well conditioned functions ($f_1$--$f_9$), particularly at lower dimensions, but faces significant challenges on highly multimodal functions ($f_{15}$--$f_{24}$) in dimensions $d \geq 5$, where it often exhausts its evaluation budget without reaching target precision. The superior performance of LQ-CMA-ES on separable functions ($f_1$--$f_5$) suggests that the global MLP surrogate does not exploit additive structure, while the superior performance of LMM-CMA-ES on selected multimodal functions suggests that local meta models are more effective than global ones on complex landscapes. The overall analysis of 24 functions in 4 dimensions shows that surrogate model selection is essential for optimization performance, and that a one size fits all MLP ensemble has inherent limitations in adapting to diverse problem structures. Future directions include adaptive surrogate selection switching between global and local models in response to detected landscape properties and improved exploration strategies for multimodal functions.

\section*{Acknowledgements}
The research reported in this paper has been supported by the Czech Academy of Science grant 231523 and Martin Holeňa’s research was partially supported by the German Research Foundation (DFG) project funding no. 441926934.

\bibliography{mybibliography}





\end{document}